\documentclass{article}

\usepackage[preprint]{tmlr}

\usepackage{amsmath,amsfonts,bm}

\def\eqref#1{equation~\ref{#1}}

\def\1{\bm{1}}

\DeclareMathAlphabet{\mathsfit}{\encodingdefault}{\sfdefault}{m}{sl}
\SetMathAlphabet{\mathsfit}{bold}{\encodingdefault}{\sfdefault}{bx}{n}

\usepackage[utf8]{inputenc} 
\usepackage[T1]{fontenc}    
\definecolor{cvprblue}{rgb}{0.21,0.49,0.74}
\usepackage[colorlinks=true,
            linkcolor=red,      
            anchorcolor=cvprblue,    
            citecolor=cvprblue,  
            pagebackref
            ]{hyperref}
\usepackage{url}            
\usepackage{booktabs}       
\usepackage{amsfonts}       
\usepackage{nicefrac}       
\usepackage{microtype}      
\usepackage{xcolor}         
\usepackage{graphicx}
\usepackage[colorinlistoftodos]{todonotes}
\usepackage{subcaption}
\usepackage{cleveref}
\usepackage{wrapfig}
\usepackage[ruled,noend,vlined]{algorithm2e}
\usepackage{etoolbox} 

\makeatletter
\patchcmd{\algocf@makecaption@ruled}{\hsize}{\textwidth}{}{} 
\patchcmd{\@algocf@start}{-1.5em}{0em}{}{} 
\makeatother

\SetAlCapHSkip{0pt} 
\usepackage{algpseudocode}
\usepackage{bm}
\usepackage{tikz}
\usepackage{fontawesome}
\usepackage{amsmath}
\usepackage{amssymb}
\usepackage{bbm}
\usepackage{pifont}
\usepackage{fancybox}
\usepackage{enumitem}
\usepackage{color, colortbl}
\usepackage{dsfont}
\usepackage{multirow} 
\usepackage[referable]{threeparttablex}
\usetikzlibrary{positioning}
\usetikzlibrary{arrows}
\usetikzlibrary{calc}
\usetikzlibrary{fit, shadows, shapes, backgrounds}
\usetikzlibrary{bayesnet}
\usepackage{xr}
\usepackage{comment}
\usepackage[toc,page,header]{appendix}
\usepackage{minitoc}
\usepackage{setspace}

\usepackage{graphicx}
\usepackage{array}
\usepackage{longtable}
\usepackage{booktabs}
\usepackage{xcolor}
\usepackage{colortbl}
\usepackage{tikz}
\usetikzlibrary{shadows}

\usetikzlibrary{shapes,decorations,arrows,calc,arrows.meta,fit,positioning}
\tikzset{
    -Latex,auto,node distance =1 cm and 1 cm,semithick,
    state/.style ={ellipse, draw, minimum width = 0.7 cm},
    point/.style = {circle, draw, inner sep=0.04cm,fill,node contents={}},
    bidirected/.style={-Latex,dashed},
    el/.style = {inner sep=2pt, align=left, sloped},
   box/.style={rectangle,draw,node distance=1cm,text width=15em,text centered,rounded corners,minimum height=2em,thick},
    arrow/.style={draw,-latex',thick,dashed},
}
\definecolor{Gray}{gray}{0.9}
\definecolor{LightCyan}{rgb}{0.8,0.9,0.9}

\definecolor{newcolor}{rgb}{.8,.349,.1}
\newcolumntype{a}{>{\columncolor{LightCyan}}c}

\renewlist{tablenotes}{enumerate}{1}
\makeatletter
\setlist[tablenotes]{label=\tnote{\alph*},ref=\alph*,itemsep=\z@,topsep=\z@skip,partopsep=\z@skip,parsep=\z@,itemindent=\z@,labelindent=\tabcolsep,labelsep=.1em,leftmargin=*,align=left,before={\scriptsize}}

\usepackage{xspace}
\newcommand{\eg}{\emph{e.g.}\xspace}
\renewcommand{\cite}{\citep}
\usepackage{wrapfig}
\usepackage{authblk}

\newcommand{\ourmethod}{\texttt{InfSplign}}

\title{\texttt{InfSplign:} Inference-Time Spatial Alignment of\\ Text-to-Image Diffusion Models}

\author{
    \noindent Sarah Rastegar\textsuperscript{\rm *1}
    \quad  Violeta Chatalbasheva\textsuperscript{\rm *1}
    \quad Sieger Falkena\textsuperscript{\rm *3}
    \quad Anuj Singh\textsuperscript{\rm 1}
    \quad Yanbo Wang\textsuperscript{\rm 1}\vspace{-0.2cm}\\
    \makebox[\textwidth]{%
    Tejas Gokhale\textsuperscript{\rm 2}
    \quad Hamid Palangi\textsuperscript{\rm †4}
    \quad Hadi Jamali-Rad\textsuperscript{\rm †1,3}
    }\\~\\
    \textsuperscript{\rm 1}Delft University of Technology, The Netherlands\\
    \textsuperscript{\rm 2}University of Maryland, Baltimore County, USA\\
    \textsuperscript{\rm 3}Shell Information Technology International\\
    \textsuperscript{\rm 4}Google\\
    \textsuperscript{\rm *,†}equal contribution
    \vspace{0.05in}\\
}

\begin{document}

\maketitle

\begin{abstract}\vspace{-0.4cm}
Text-to-image (T2I) diffusion models generate high-quality images but often fail to capture the spatial relations specified in text prompts. This limitation can be traced to two factors: lack of fine-grained spatial supervision in training data and inability of text embeddings to encode spatial semantics. We introduce \texttt{InfSplign}, a training-free inference-time method that improves spatial alignment by adjusting the noise through a compound loss in every denoising step. Proposed loss leverages different levels of cross-attention maps extracted from the backbone decoder to enforce accurate object placement and a balanced object presence during sampling. The method is lightweight, plug-and-play, and compatible with any diffusion backbone. Our comprehensive evaluations on VISOR and T2I-CompBench show that \texttt{InfSplign} establishes a new state-of-the-art (to the best of our knowledge), achieving substantial performance gains over the strongest existing inference-time baselines and even outperforming the fine-tuning-based methods. Codebase is available at \href{https://github.com/VioletaChatalbasheva/InfSplign}{GitHub}.
\end{abstract}

\section{Introduction}
\begin{figure}
  \centering
  \includegraphics[width=0.48\linewidth]{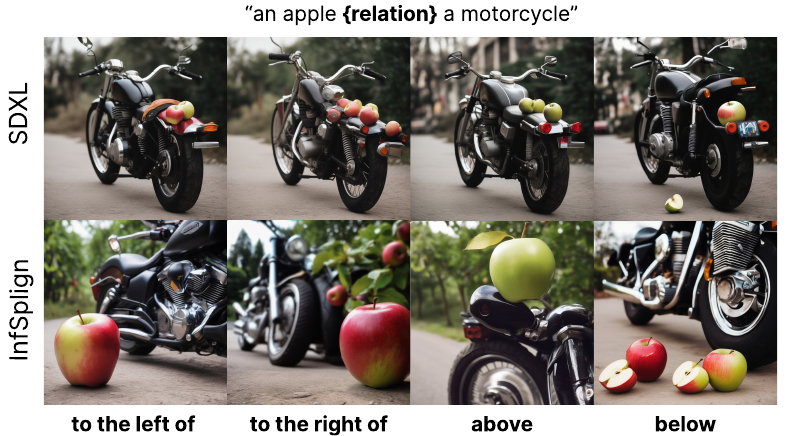}
  \includegraphics[width=0.48\linewidth]{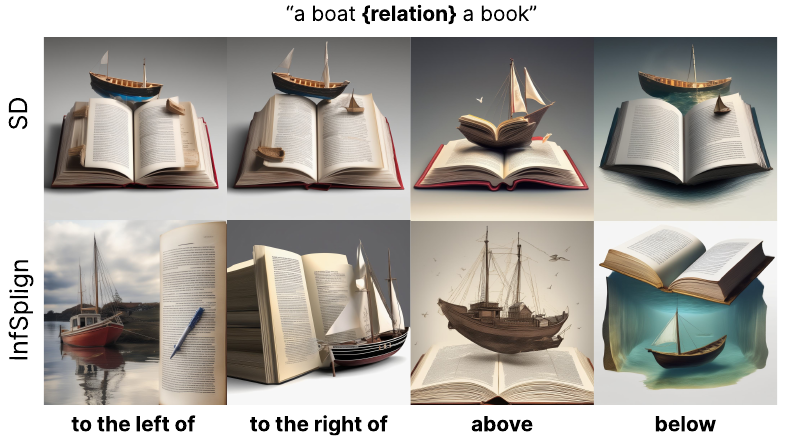}\vspace{-0.3cm}
  \caption{
  \ourmethod{} is a training-free inference-time method that improves spatial understanding of text-to-image (T2I) Stable Diffusion (SD) models, namely SD v$1.4$, SD v$2.1$ and SDXL.
  }
  \label{fig:fig1}
\end{figure}

\vspace{-0.3cm}
Diffusion-based text-to-image (T2I) generative models have rapidly advanced, enabling the synthesis of high-quality, detailed images from arbitrary textual descriptions~\cite{rombach2022high,saharia2022photorealistic,Dhariwal2021DiffusionSynthesis}. 
Despite these developments, precise control over spatial relationships described in text prompts remains challenging, manifesting as misplacement or unintended merging of objects in generated images or even completely failing to depict all specified objects or attributes~\cite{visor}.
Diffusion models frequently fail to distinguish between prompts such as \textit{"object A to the left of object B"} and \textit{"object A to the right of object B"}, often producing nearly identical outputs irrespective of spatial cues as shown in \autoref{fig:fig1}. This misalignment substantially reduces reliability, hindering applications that demand accurate spatial reasoning, such as generating scene layouts for robotic manipulation and visual grounding in augmented reality systems~\cite{Chen_2024_CVPR}. Beyond misplacement, a key limitation is that cross-attention maps are often misaligned with the true object regions, yielding inaccurate centroid estimates and consequently, reduced spatial accuracy. The deficiency in spatial understanding is quantitatively evident; for instance, on the T2I-CompBench~\cite{huang2023t2i} compositional reasoning benchmark, state-of-the-art performances on spatial understanding are around $20\%$, significantly lagging behind performance on other aspects such as attribute binding (around $50\%$).
This performance gap underscores a critical area within T2I research, emphasizing the importance of developing solutions that effectively address spatial misalignment. 

A common hypothesis is that these limitations in spatial cognizance stem from the BERT-style CLIP text encoder \cite{radford2021learning}, trained with a global image-text contrastive objective and no explicit spatial supervision, fails to learn sufficiently precise spatial relations \cite{visor}. Other studies have hypothesized that classifier-free guidance (CFG) might be entangling multiple semantic factors in the text prompt \cite{wu2024contrastive}, and that ``positive'' prompt might be too weak to enforce spatial alignment \cite{chang2024contrastiveCFG}, proposing a contrastive setting with both positive and negative prompts. Another line of research investigates the impact of the amount of training data exhibiting spatial relationships on model performance; for \eg, SPRIGHT~\cite{chatterjee2024getting} generates new spatially-focused captions for four widely-used datasets and finetunes the model on them.

Approaches tackling spatial accuracy or object preservation broadly fall into two categories: \emph{fine-tuning-based} and \emph{inference-time} methods. Fine-tuning-based methods typically employ spatially-aware datasets \cite{zhang2024compass}, auxiliary reward models \cite{zhang2024itercomp}, or explicit training mechanisms \cite{clark2024directly}, achieving higher spatial accuracy at the cost of considerable computational overhead and the risk of negatively impacting the carefully optimized diffusion backbone and its generalizability. In contrast, inference-time methods~\cite{voynov2023sketch,epstein2023diffusion} avoid expensive re-training, providing computationally-efficient alternatives capable of flexible spatial adjustments during sampling. Some of the existing inference-time methods remain to be overly complex, relying on auxiliary inputs like layout maps~\cite{sun2024spatial}, scene graphs~\cite{farshad2023scenegenie}, or external guidance from large language models (LLMs)~\cite{phung2024grounded,lian2024llmgrounded}, thereby limiting ease of deployment and interoperability.

We focus on inference-time approaches and directly extract spatial information from the U-Net decoder attention maps during the reverse diffusion to guide the sampling process. Building on the insight that attention maps can act as proxies for spatial information \cite{han2025spatial, epstein2023diffusion}, we introduce \texttt{InfSplign}, an inference-time method that partitions the U-Net decoder cross-attention maps into three hierarchical levels: coarse, mid-level, and fine-grained. From the coarse and mid-level attentions, we extract object centroids and variances, which are then used to define three complementary loss terms. Together, these losses enforce spatial alignment and promote balanced object representations in the final denoised outputs. At each denoising step, the compound loss is applied to refine the predicted noise for the subsequent timestep, thereby guiding the sampling trajectory toward spatially coherent images. 

\textbf{Remarks.} Widely-adopted spatial benchmarks concentrate mainly on four primary object relationships (i.e., \textit{left}, \textit{right}, \textit{top}, and \textit{bottom}), and existing baselines reveal a substantial performance gap even for addressing these fundamental cases. Consequently, we focus on these relationships. Our core contributions are as follows:
\begin{itemize}[nosep,leftmargin=*]
\item[--] \textit{Spatial Alignment and Object Preservation.} We introduce \texttt{InfSplign}, a training-free inference-time approach that leverages attention maps to enforce object spatial alignment and preservation. Our proposed compound loss is comprised of three components: an object location loss that enforces accurate spatial grounding, an object presence loss, which increases the certainty of object representation, and an object balance loss, which mitigates cross-object interference in the finer layers of the U-Net decoder.

\item[--] \textit{Experimental Results.} Extensive evaluations on spatial benchmarks, including VISOR and T2I-CompBench, demonstrate that \texttt{InfSplign} improves spatial alignment by up to $24.81\%$ and $21.91\%$ over state-of-the-art inference-time methods, and even surpasses fine-tuning approaches by $14.33\%$ and $9.72\%$. Extensive ablation studies and qualitative results further corroborate the effectiveness of our proposed approach. 

\end{itemize}

\section{Related Work}\vspace{-0.2cm}
\textbf{Text-to-Image (T2I) Generation} aims to produce visually realistic images that align with natural language prompts. While earlier work focused on GANs~\cite{isola2017image, li2022blip, park2019semantic} and autoregressive models~\cite{ramesh2021zero, yu2022scaling}, diffusion models~\cite{ho2020denoising, rombach2022high, saharia2022photorealistic} have become the dominant approach due to their superior image fidelity, diversity, and stability~\cite{liu2022compositional}. The integration of vision–language pretraining strategies such as CLIP~\cite{radford2021learning} helped further enhance semantic alignment~\cite{ramesh2022hierarchical}, yet recent studies show that even state-of-the-art models struggle with accurately capturing fine-grained textual details, particularly spatial relationships~\cite{huang2023t2i, wang2024information}. To address this, expansion of network architectures or new training objectives have been explored~\cite{feng2023trainingfree, li2024matting}, but these require costly retraining. 

\textbf{Spatial Understanding and Object Preservation in T2I Models.}
Efforts aimed at improving spatial understanding primarily fall into fine-tuning-based or layout-conditioned methods. 
Fine-tuning approaches enhance spatial reasoning by training models on spatially-aware datasets or using auxiliary objectives, such as reward-based optimization~\cite{zhang2024large, chen2023learning, zhang2024itercomp}.
SPRIGHT \cite{chatterjee2024getting} presents a large scale vision-language dataset for fine-tuning diffusion model for spatial data. CoMPaSS~\cite{zhang2024compass} advances spatial accuracy by explicitly incorporating spatially labeled data during training. 
However, these methods involve expensive retraining processes and risk destabilizing the pretrained diffusion backbone. 
Another category explicitly injects spatial layout information (sometimes through LLMs), such as bounding boxes, depth maps, or segmentation masks, to guide generation~\cite{sun2024spatial, chen2024training, gong2024check, phung2024grounded, li2023gligen, lee2024reground, nie2024compositional, lian2024llmgrounded, derakhshani2023unlocking}. 
However, these methods depend on external layout inputs which may not always be available, thus limiting usability, and require additional pre-processing and computational overhead.

\textbf{Inference-Time Guidance for Diffusion Models} circumvent costly retraining by directly manipulating diffusion processes during sampling. Methods like Attend \& Excite~\cite{chefer2023attend} address the issue of missing objects by optimizing attention maps at inference time, but do not explicitly enforce spatial accuracy. Structured Diffusion Guidance~\cite{feng2023trainingfree} manipulates attention maps for improved layout control using additional text preprocessing techniques, yet lacks explicit modeling of spatial relationships described by textual prompts.
Composable Diffusion~\cite{liu2022compositional} interprets diffusion models as energy-based compositions of individual concepts, improving object presence but providing minimal spatial control. More targeted spatial inference-time methods, such as Prompt-to-Prompt~\cite{hertz2023prompttoprompt} and DIVIDE\&BIND~\cite{DivideAndBind}, demonstrate the potential of directly modifying cross-attention maps. Recent information-theoretic insights further motivate inference-time interventions: analysis of mutual information between text prompts and images \cite{wang2024information, kong2024interpretable}, initial diffusion noise predetermines object layout generation \cite{ban2024crystal}, and sometimes needs to be guided to produce a valid sample \cite{guo2024initno} according to the prompt. Diffusion Self-Guidance~\cite{epstein2023diffusion} develops a framework for image editing by controlling the appearance, shape, size and location of objects but limits sample diversity. 
REVISION~\cite{chatterjee2024revision} generates spatially accurate synthetic images as conditional input, reducing the task to an image-to-image (I2I) pipeline. 
STORM~\cite{han2025spatial} introduces a distribution-based loss using Optimal Transport (OT)~\cite{villani2008optimal} to adjust attention maps toward target distributions fixed at specified spatial relationships. 
Unlike prior approaches which rely on external targets or synthetic images, \ourmethod{} directly regulates object preservation during sampling, ensuring both alignment and completeness.

\section{\texttt{InfSplign}: \underline{Inf}erence-time 
\underline{Sp}atial A\underline{lign}ment}

\subsection{Preliminaries}

\textbf{Diffusion Models} provide an effective framework for sampling from complex data distributions $q(x)$ by learning to invert a forward diffusion process. 
The forward process is a Markov chain that iteratively adds Gaussian noise to a clean data point $x_0 \in \mathcal{X}$ over $T$ steps: $x_t = \sqrt{\bar\alpha_t}\,x_0 + \sqrt{1-\bar\alpha_t}\,\epsilon_t$, where $\epsilon_t \sim \mathcal{N}(0,I)$, and $\{\beta_t\}_{t=1}^T$ is a variance schedule with $\alpha_t = 1-\beta_t$ and $\bar\alpha_t = \prod_{i=1}^t \alpha_i$~\citep{ho2020denoising,nichol2021improved}.
The reverse process defines a generative model $p_\theta(x_{t-1}|x_t)$ that approximates the true posterior $q(x_{t-1}|x_t, x_0)$. 
A neural network $\epsilon_\theta(x_t, t)$, typically a U-Net~\citep{ronneberger2015unet}, is trained to predict the noise added during the forward process. At inference time, a simplified update rule can be written as $x_{t-1} \approx x_t - s_t \,\epsilon_\theta(x_t, t)$, where $s_t$ is a step-size factor depending on the variance schedule. Incorporating conditioning variable $y$ (in our case, text prompts) results in conditional predictions $\epsilon_\theta(x_t, t, y)$~\citep{zhang2023adding,Mo2023FreeControl:Condition}. Our work is built upon a text-conditioned latent diffusion model, Stable Diffusion \citep{rombach2022high}, 
which operates in latent space $z_t$ obtained from a pretrained variational autoencoder.


\textbf{Inference-Time Guidance of Diffusion Models.}
Diffusion models can be adapted to a wide range of downstream tasks at inference time (without retraining or fine-tuning) through input conditioning (\eg text prompts) and external reward modes (\eg CLIP-based scores), to influence the denoising strategy to better align with desired outcomes.
Classifier guidance \cite{Dhariwal2021DiffusionSynthesis} steers generation using gradients from a pretrained image classifier, whereas classifier-free guidance \cite{Ho2021Classifier-FreeGuidance} eliminates the need for an external classifier by training the model to denoise both with and without conditioning, and then interpolating between the two at inference time. 
Diffusion models can also be interpreted through their score-based formulation, where the model estimates the gradient of the log probability density, $\nabla_{z_t} \text{log} p(z_t, t)$. This gives us an intuition about the direction to move in to increase the log likelihood of our data sample based on some conditional information. While the denoising formulation, $\epsilon_\theta (z_t, t)$, gives us a prediction of the noise that was added by the forward diffusion model at each timestep, the inference-time classifier-free guidance (CFG) approach~\citep{Ho2021Classifier-FreeGuidance} is commonly adopted to guide a conditional reverse diffusion process toward a desired conditioning signal (a text prompt $y$ in our T2I setting). The score-based formulation of CFG is created from two conditional and unconditional terms:
\begin{equation}
    \label{eq:cfg_score}
    \nabla_{z_t} \text{log } p(z_t | y, t) \approx \nabla_{z_t} \text{log } p(z_t, t) + \gamma (\nabla_{z_t} \text{log } p(z_t | y, t) - \nabla_{x_t} \text{log } p(z_t, t)),
\end{equation}
where $\gamma$ is the guidance strength. The equivalent noise prediction form at step $t$ can be given by:
\begin{equation}
\epsilon_t \;=\; \epsilon_\theta(z_t; t) 
\;+\; \gamma ( \epsilon_\theta(z_t; t, y) - \epsilon_\theta(z_t; t) ).
\end{equation}
%

%
%


%
\begin{wrapfigure}{r}{0.5\textwidth}
  \centering
  \vspace{-2\intextsep}
  \includegraphics[width=1\linewidth]{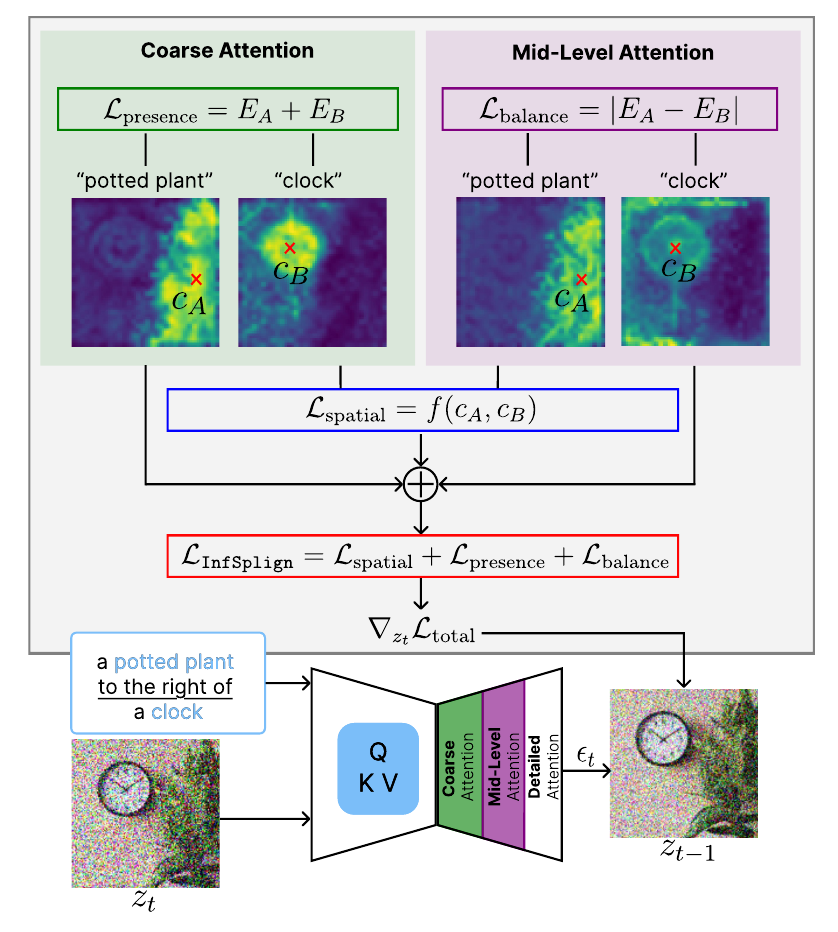}\vspace{-0.3cm}
  \caption{
  Overview of the proposed approach.} 
  \label{fig:pipeline_losses}
\end{wrapfigure}
\subsection{A Deeper Dive into \ourmethod{}.} 

Our work is focused on inference-time guidance, and serves as a lightweight, plug-and-play enhancement to diffusion models. 
To tackle with data and caption limitations, we introduce a guidance signal to quantify the misalignment between the generated latent attention maps and the spatial cues in the prompt, as well as to ensure a balanced representation of all objects throughout the reverse diffusion process. The idea is to \emph{actively nudge} the generation process towards generating more spatially-cognizant images. 
\autoref{fig:pipeline_losses} gives a high-level overview of the mechanics of \ourmethod{}. The input to the system is a user-specified text prompt, \eg \textit{"a potted plant to the right of a clock"}, and the noisy latent embedding $z_t$ produced at timestep $t$ of the reverse diffusion process. We represent the prompt as a structured triplet - $\langle A, R, B \rangle$, where $A$ and $B$ are the object tokens and $R \in \mathcal{R}$, where $\mathcal{R}$ is the set of spatial relationships, \{\text{"left"}, \text{"right"}, \text{"above"}, \text{"below"}\}, e.g. $A{=}\text{"potted plant"}$, $B{=}\text{``clock''}$ and $R{=}\text{``to the right of''}$.

To guide the denoising process, we extract the cross-attention maps corresponding to the object tokens $A$ and $B$ at timestep $t$. 
We divide the attention maps into three abstraction levels: 
(i) \textbf{coarse attention:} where presence of the objects is shaped; 
(ii) \textbf{mid-level attention:} at which we observed objects might dominate each other in magnitude as such impacting their representation in the final outcome; 
(iii) \textbf{fine-grained attention:} where nuances and high-resolution subtleties of the final image appear. 
We focus on the first two levels (coarse and mid-level), where spatial cognizance of the model is formed. 
As we elaborate later on, our loss is comprised of three terms: \textbf{spatial alignment loss} ($\mathcal{L}_{\textup{spatial}}$), \textbf{object presence loss} ($\mathcal{L}_{\text{presence}}$), \textbf{representation balance loss} ($\mathcal{L}_{\text{balance}}$) constituting our proposed total loss $\mathcal{L}_{\texttt{InfSplign}}$. Both coarse and mid-level attention maps are used to estimate spatial alignment loss. We utilize the coarse-level attention layers for the presence loss to minimize the variance of each object's attention map, thus ensuring that all objects will be maintained through the reverse diffusion steps. We use mid-level attention layers in the balance loss to ensure that one object does not dominate the other. The gradient of $\mathcal{L}_{\texttt{InfSplign}}$ with respect to (w.r.t.) the latent $z_t$ provides a guidance signal that modifies the predicted noise, shifting the latent toward improved spatial alignment:
%
\begin{equation}
    \label{eq:spatial_cfg}
    \epsilon_{t} \gets \epsilon_{\theta}(z_t;t) + \gamma (\epsilon_{\theta}(z_t;t,y) - \epsilon_{\theta}(z_t;t)) + \eta \nabla_{z_t}\mathcal{L_\texttt{InfSplign}} = \epsilon_{t} + \nabla_{z_t}\mathcal{L_\texttt{InfSplign}},
\end{equation}
where $\eta$ is a weight parameter acting similar to guidance strength ($\gamma$), balancing the magnitude of the last two terms. This updated noise prediction is used to compute the denoised latent $z_{t-1}$ guiding the generation towards spatially cognizant images. Thus, the final update direction combines both semantic (via CFG) and overall spatial alignment (imposed via $\mathcal{L}_{\texttt{InfSplign}}$). This is applied iteratively over all reverse timesteps. 

\textbf{From Attention to Centroids and Variances.} It is demonstrated in
\cite{hertz2023prompttoprompt} that cross-attention layers encode rich information about the spatial arrangement of objects in generated images. Building on this insight, we extract attention maps from coarse- and mid-level decoder layers of the U-Net, 
which most reliably capture object structure and spatial location. We estimate the position of each object by computing the centroid of its attention distribution. The centroid $c_A$ of token $A$ is computed as a weighted average over the spatial coordinates of the latent $z_t$, using the attention layer $l$ weights of token $A$ (denoted by $\mathcal{A}_t^{(l)}$) as coefficients and normalized by the total attention mass: 
\begin{equation}
\label{eq:centroid}
c_A^{(l)} = (x_A, y_A) =
\left(
\frac{\sum_{h,w} \mathcal{A}_t^{(l)}[h, w] \cdot x_w}
     {\sum_{h,w}  \mathcal{A}_t^{(l)}[h, w]},
\quad
\frac{\sum_{h,w} \mathcal{A}_t^{(l)}[h, w] \cdot y_h}
     {\sum_{h,w}  \mathcal{A}_t^{(l)}[h, w]}
\right),
\end{equation}
where $h \in [H]$ and $w \in [W]$ scan over the height and width of the corresponding attention map of size $H \times W$, where the origin, $(1,1)$, is assumed to be on the top-left of the tensor. Note that since the centroids are derived from U-Net cross-attention maps, which themselves depend on the latent representation $z_t$, the resulting loss function in \autoref{eq:final_loss} remains differentiable w.r.t. $z_t$. Additionally, we incorporate variance as a measure of uncertainty in the attention maps. Low variance indicates high attention values close to the centroid and low attention away from it, thus encouraging distinct object representations, while high variance suggests weak attention, risking object omission. So, we compute the variance $\sigma^{2(l)}_A$ of the attention layer $l$ distribution of token $A$ by weighting the squared distance of each pixel from the centroid $c_A$ (in \autoref{eq:centroid}) by its attention value, normalized by the total attention mass:
%
\begin{equation}
\label{eq:varA}
\sigma_A^{2(l)} =
\left(
\frac{\sum_{h,w} \mathcal{A}_t^{(l)}[h, w] \cdot {\parallel}(x_w,y_h)-c_A^{(l)}{\parallel}^2}
     {\sum_{h,w}  \mathcal{A}_t^{(l)}[h, w]}
     \right).
\end{equation}
%
%
\textbf{Spatial Alignment Loss.} 
The spatial relationship $R$ between two objects $A$ and $B$ is expressed as a difference between their centroids, denoted as $\Delta$, and is computed along the appropriate axis as:
%
\begin{equation}
\label{eq:Delta}
\Delta = \left\{
\begin{alignedat}{2}
x_B - x_A &\ \text{ for \textit{"left"}},  &\quad x_A - x_B &\ \text{ for \textit{"right"}},\\
y_B - y_A &\ \text{ for \textit{"above"}}, &\quad y_A - y_B &\ \text{ for \textit{"below"}},\quad \|x_A - x_B\| \ \text{ for \textit{"near"}}.
\end{alignedat}
\right.
\end{equation}
This captures the directional alignment between the two objects and signals adherence to the specified spatial relation. Building on this, we define the spatial loss function as:
\begin{equation}
    \label{eq:spatial_loss}
    \mathcal{L_\text{spatial}} =  f_\text{spatial}(\alpha(m - \Delta)),
\end{equation}
where $f_{\text{spatial}}$ denotes a pointwise activation-based loss which allows us to penalize spatial inconsistencies in a controlled manner, $m$ is a distance margin indicating the acceptable minimum distance between the objects' centroids, and $\alpha$ is a scaling factor controlling the steepness of the loss. Notably, $\alpha$ sharpens the slope around the decision boundary, determining how strictly the model is penalized. $\mathcal{L_\text{spatial}}$ results in a small penalty if the objects are placed correctly w.r.t.\ the target spatial relation, \eg if $\mathcal{S}={\text{\textit{``to the right of''}}}, x_A {-} x_B > m, \mathcal{L_{\text{spatial}}} \to 0$. The loss is high when objects violate spatial relation, are too close to each other, or $\Delta < m$. 



\begin{figure}[t!]
  \centering
  \includegraphics[width=0.9\linewidth]{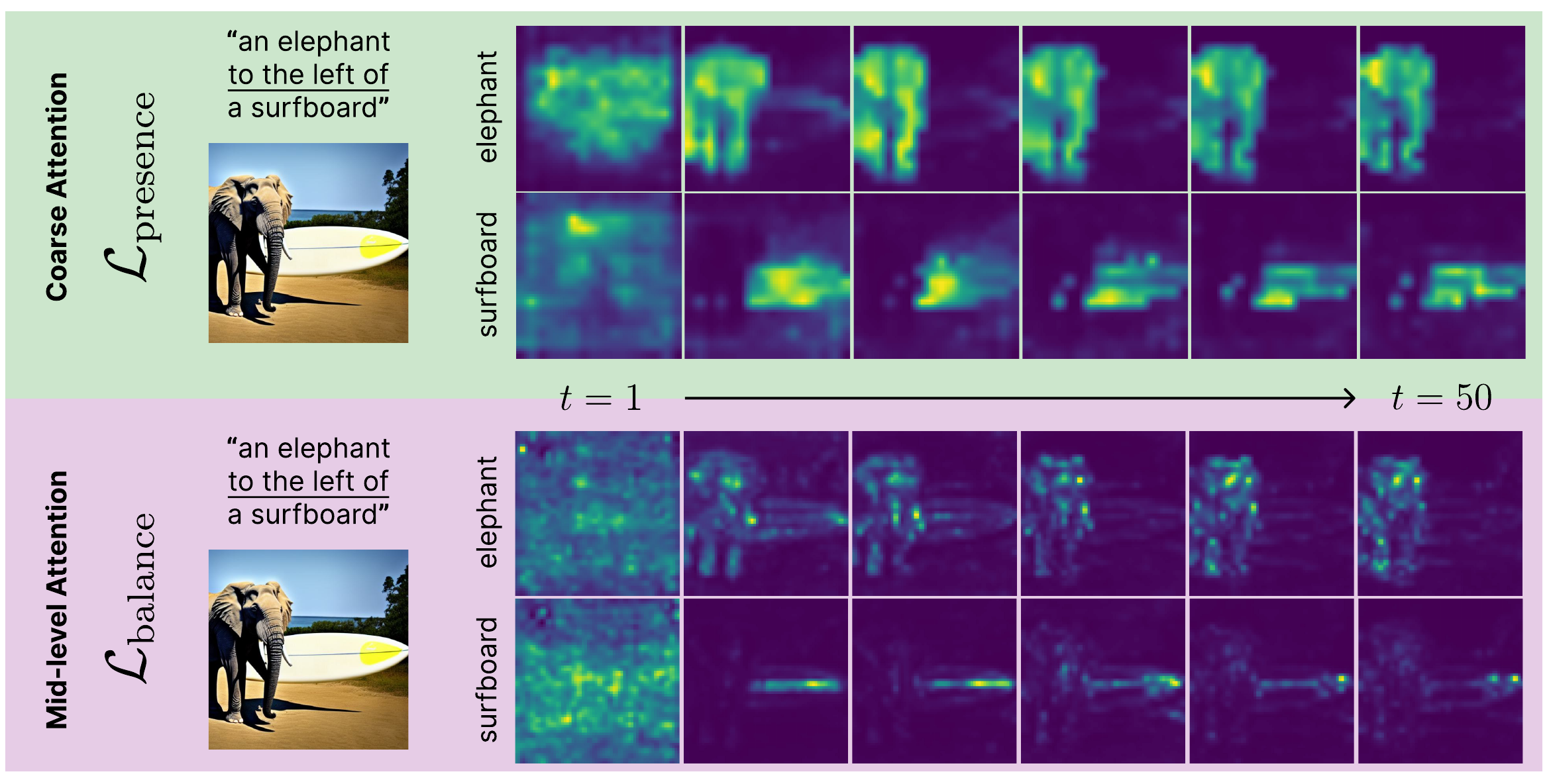}
  \caption{Attention energy across decoder cross-attention layers. Coarse layers encode global structure, with $\mathcal{L}_{\text{presence}}$ enforcing focused attention; mid-level layers encode local detail, with $\mathcal{L}_{\text{balance}}$ equalizing object energy to prevent dominance.} 
  \label{fig:attn_presence_balance}
\end{figure}

\textbf{Object Presence Loss.}
A prerequisite of spatial alignment is to ensure both objects remain visible in the final denoised image. 
We examine how widely the attention for each token is spread across the image. As shown in \autoref{fig:attn_presence_balance}, attention often becomes more focused over time. Pixels near the centroid tend to carry stronger attention than those farther away. A smaller variance means the model has a clearer, more focused idea of where the object is. When the variance is large and the attention is spread out, the model is uncertain about the object’s location, making it more likely to appear weak or disappear in the final image. To address this, we introduce an object presence loss, $\mathcal{L}_{\text{presence}}=\sigma_A^{2(l)}+\sigma_B^{2(l)}$, which reduces the variance of each object’s attention map in layer $l$. Minimizing this variance encourages the attention to concentrate more tightly around the object.

\textbf{Representation Balance Loss.}
Another common failure is that one object is omitted when its attention is much weaker than the other's, $\sigma_A \gg \sigma_B$. At coarse layers this imbalance matters less, but at mid-level layers the weaker object can be suppressed. To mitigate this, we enforce that both objects maintain similar levels of magnitude. As shown in \autoref{fig:attn_presence_balance}, these coarse attention maps often capture clusters of fine-grained details. In such layers, simply minimizing variance would undesirably restrict the model’s ability to explore object details. Instead, we ensure parity of uncertainty across objects by introducing an object balance loss, $\mathcal{L}_{\text{balance}}=|\sigma_A^{2(l)}-\sigma_B^{2(l)}|$, which encourages both objects to have a similar degree of dispersion in their attention maps in layer $l$. This prevents one object from overshadowing the other, ensuring a balanced representation. We arrive at our proposed spatial alignment loss (which we refer as $\mathcal{L}_\texttt{InfSplign}$) by jointly enforcing correct placement, object preservation and a balanced representation of objects, with hyperparameters ${\lambda_{\text{s}}}$, ${\lambda_{\text{p}}}$, and ${\lambda_{\text{b}}}$ controlling the relative strength of each component:
\begin{equation}
    \label{eq:final_loss}
    \mathcal{L_\text{\texttt{InfSplign}}} = {\lambda_\text{s}}\mathcal{L_\text{spatial}} +{\lambda_\text{p}} 
    \mathcal{L_\text{presence}} + {\lambda_\text{b}}\mathcal{L_\text{balance}}. 
\end{equation}

\begin{wrapfigure}{r}{0.5\linewidth}
    \LinesNumbered
    \IncMargin{1em} 
    \begin{algorithm}[H]
        \small
        \SetKwInput{Require}{Require}
        \SetKwInput{Return}{Return}
        \SetAlgoLined
        \DontPrintSemicolon
        \SetNoFillComment
     
        \Require{$\mathcal{P} = \langle A, R, B \rangle$, $z_t$, \texttt{SD}(.), $\eta$}

        $\mathcal{A}_t, \mathcal{B}_t, \epsilon_t \gets \texttt{SD}(z_t, \mathcal{P})$ 
        
        $c_A, c_B \gets \texttt{Centroid}(\mathcal{A}_t, \mathcal{B}_t)$ using \autoref{eq:centroid}

        $\sigma^2_A, \sigma^2_B \gets \texttt{Variance}(\mathcal{A}_t, \mathcal{B}_t, c_A, c_B)$ using \autoref{eq:varA}

        $\Delta \gets \texttt{Difference}(c_A, c_B, R)$ using \autoref{eq:Delta}

        $\mathcal{L}_{\text{spatial}} \gets f_{\text{spatial}}(\Delta, \alpha, m)$ using \autoref{eq:spatial_loss}
        
        $\mathcal{L}_{\text{presence}} \gets \sigma^2_A+\sigma^2_B$, $\mathcal{L}_{\text{balance}} \gets |\sigma^2_A-\sigma^2_B|$
        
        $\mathcal{L_\texttt{InfSplign}} = {\lambda_\text{s}}\mathcal{L_\text{spatial}} +{\lambda_\text{p}}    \mathcal{L_\text{presence}} + {\lambda_\text{b}}\mathcal{L_\text{balance}}$
         
        $\epsilon_t \gets \epsilon_t + \eta \cdot \nabla_{z_t} \mathcal{L}_\texttt{InfSplign}$ using \autoref{eq:spatial_cfg}
        
        $z_{t-1} \gets z_t - s_t \cdot \epsilon_t$ 
        

         
        

    
        \Return{$z_{t-1}$}
        
        \caption{Denoising with \ourmethod{}}\label{alg:pseudocode}
    \end{algorithm}
\vspace{-5em}
\end{wrapfigure}

Unlike Attend-and-Excite \cite{chefer2023attend}, which implicitly enforces object presence by maximizing attention weights of the neglected tokens or STORM \cite{han2025spatial}, which fixes the object locations to achieve adherence with the spatial cues in the text prompt, our proposed method preserves generation diversity by penalizing misplaced or omitted objects through gradients computed w.r.t the latent $z_t$ during sampling. 
%
%
The pseudocode for a single reverse diffision step under \ourmethod{} is in \autoref{alg:pseudocode}.

\section{Experiments}
We present a comprehensive evaluation showing the efficacy of our approach against state-of-the-art baselines. Next, we conduct detailed ablation studies to assess the impact of the key (hyper)parameters. Finally, we present qualitative results that highlight the superior performance of \ourmethod. 

\textbf{Implementation Details.} 
We apply \texttt{InfSplign} on top of SD v$1.4$ and v$2.1$ models \cite{rombach2022high} for $50$ inference steps, which is standard protocol adopted by prior work~\citep{han2025spatial, chatterjee2024revision, chefer2023attend, feng2023trainingfree}. The hyperparameters $\alpha$ and margin are set to $\alpha = 1.5$, $m = 0.25$ (SD v1.4), $m = 0.5$ (SD v2.1) respectively, and $f_{\text{spatial}}(.)$ to \texttt{GeLU} \citep{hendrycks2016gaussian}, CFG guidance scale $\gamma = 7.5$, and the guidance weight for our loss is set to $\eta = 1000$ to balance magnitudes. For \emph{coarse attention} and \emph{mid-level attention}, we use the cross-attention layers $1\!-\!3$ of the first and second blocks of the U-Net decoder, respectively. The $\mathcal{L}_{\text{presence}}$ loss is computed from the attention maps of the first block, while $\mathcal{L}_{\text{balance}}$ is derived from those of the second block. Hyperparameters are selected via grid search (see \autoref{app:extend_ablations}), yielding $\lambda_{\text{s}}{=}0.5$, $\lambda_{\text{p}}{=}1$, $\lambda_{\text{b}}{=}0.5$ for SD v1.4 and $\lambda_{\text{s}}{=}0.5$, $\lambda_{\text{p}}{=}1$, $\lambda_{\text{b}}{=}1.0$ for SD v2.1. We use the same random seeds for each benchmark as defined in the original papers.

\textbf{Baselines and Benchmarks.} We compare our approach with the most relevant and recent baselines, both \emph{inference-time} \cite{han2025spatial,chatterjee2024revision,chefer2023attend,feng2023trainingfree,DivideAndBind,guo2024initno,meral2024conform} and \emph{fine-tuning} \cite{chatterjee2024getting, zhang2024compass,liu2022compositional}. We evaluate our performance on the two most commonly adopted benchmarks assessing spatial understanding of diffusion models, namely VISOR \cite{visor} and T2I-CompBench \cite{huang2023t2i}. Notably, VISOR is focused on evaluating how well T2I models generate correct spatial relationships described in the text prompt, whereas T2I-CompBench assesses T2I models on broader compositional metrics such as attribute binding, numeracy, and complex compositions, with a module dedicated to object relations pertinent to our evaluation angle.    

\textbf{Quantitative Evaluation: VISOR.} \autoref{tab:visor_results} summarizes the VISOR scores of our method compared to relevant approaches in T2I space. As can be seen, \texttt{InfSplign} consistently outperforms both fine-tuning and inference-time methods across all baselines by a significant margin. On SD v$1.4$, \ourmethod~surpasses the best of inference-time methods (such as CONFORM, INITNO, and STORM), as well as fine-tuning-based methods (e.g., SPRIGHT and CoMPaSS). We achieve the highest score across all metrics. Most notably on object accuracy (OA), we outperform the best of inference-time and fine-tuning based competitors by $6.35\%$ and $1.8\%$, respectively. The same holds for the most challenging task of the benchmark, i.e. "unconditional" score, with $8.96\%$ and $9.13\%$, and VISOR-$4$, with $9.39\%$ and $7.63\%$, respectively. This demonstrates that \texttt{InfSplign} can surpass the performance of models that require additional input and/or retraining. These results indicate that when both objects are successfully generated, our method ensures their spatial relationship is correct. This trend continues to further improve with the stronger backbones SD v$2.1$ consistently. The margin is significantly larger, where we report up to $24.81\%$ and $14.33\%$ score improvement on the challenging VISOR-$4$ task compared to the state-of-the-art inference-time and fine-tuning baselines, STORM and ComPaSS, respectively. This shows that our method reliably produces four spatially correct images for every test prompt. One key factor behind these substantial improvements is the higher quality of attention maps with stronger backbones, as our method relies on attention maps to guide the denoising process.

\begin{table*}[t!]
    \caption{
        \textbf{Performance comparison between different models on VISOR (\%) and Object Accuracy (OA) (\%) metrics}, 
        based on Stable Diffusion 1.4, and 2.1. \ding{168} reported from \cite{han2025spatial}
    }
    \Large
    \centering
    \begin{threeparttable}
        \resizebox{\linewidth}{!}{
            \begin{tabular}{lcccccccccc}
                \toprule
                \multirow{2}{*}{Model} 
                    & \multirow{2}{*}{Venue}
                    & \multirow{2}{*}{\shortstack{Fine-\\tuning}} 
                    & \multirow{2}{*}{\shortstack{Extra\\ Input}} 
                    & \multirow{2}{*}{OA (\%)} 
                    & \multicolumn{6}{c}{\textbf{VISOR (\%)}} \\
                \cmidrule(lr){6-11}
                &&&&& uncond & cond & 1 & 2 & 3 & 4 \\
                \bottomrule
                \arrayrulecolor[HTML]{dedee0}
                \rowcolor[HTML]{dedee0} \textbf{Stable Diffusion 1.4} &&&&&&&&&& \\

                \rowcolor[HTML]{FAFAFA} SD 1.4 \cite{rombach2022high} 
                    && \ding{55} & \ding{55} & 29.86 & 18.81 & 62.98 & 46.60 & 20.11 & 6.89 & 1.63 \\
                \rowcolor[HTML]{FAFAFA}SD 1.4 + CDM \cite{liu2022compositional} 
                    & {\footnotesize\textcolor{gray}{ECCV22}} 
                    & \ding{51} & \ding{55} & 23.27 & 14.99 & 64.41 & 39.44 & 14.56 & 4.84 & 1.12 \\
                \rowcolor[HTML]{FAFAFA}GLIDE \cite{pmlr-v162-nichol22a} 
                    & {\footnotesize\textcolor{gray}{ICML22}} 
                    & \ding{51} & \ding{55} & 3.36 & 1.98 & 59.06 & 6.72 & 1.02 & 0.17 & 0.03 \\
                \rowcolor[HTML]{FAFAFA}GLIDE + CDM \cite{liu2022compositional} 
                    & {\footnotesize\textcolor{gray}{ECCV22}} 
                    & \ding{51} & \ding{55} & 10.17 & 6.43 & 63.21 & 20.07 & 4.69 & 0.83 & 0.11 \\
                \rowcolor[HTML]{FAFAFA}Control-GPT \cite{zhang2023controllable} 
                    & {\footnotesize\textcolor{gray}{arXiv23}} 
                    & \ding{51} & \ding{51} & 48.33 & 44.17 & 65.97 & 69.80 & 51.20 & 35.67 & 20.48 \\
                \rowcolor[HTML]{FAFAFA}CoMPaSS \cite{zhang2024compass} 
                    & {\footnotesize\textcolor{gray}{ICCV25}} 
                    & \ding{51} & \ding{55} & 65.56 & 57.41 & 87.58 & 83.23 & 67.53 & 49.99 & 28.91 \\
                \midrule

                \rowcolor[HTML]{FAFAFA}Structure Diffusion \cite{feng2023trainingfree} 
                    & {\footnotesize\textcolor{gray}{ICLR23}} 
                    & \ding{55} & \ding{55} & 28.65 & 17.87 & 62.36 & 44.70 & 18.73 & 6.57 & 1.46 \\
                \rowcolor[HTML]{FAFAFA}Attend-and-Excite \cite{chefer2023attend} 
                    & {\footnotesize\textcolor{gray}{SIGGRAPH23}} 
                    & \ding{55} & \ding{55} & 42.07 & 25.75 & 61.21 & 49.29 & 19.33 & 4.56 & 0.08 \\
                \rowcolor[HTML]{FAFAFA}Divide-and-Bind\ding{168} \cite{DivideAndBind} 
                    & {\footnotesize\textcolor{gray}{BMVC24}} 
                    & \ding{55} & \ding{55} & 46.03 & 31.62 & 68.70 & 64.72 & 37.82 & 18.64 & 5.30 \\
                \rowcolor[HTML]{FAFAFA}INITNO\ding{168} \cite{guo2024initno} 
                    & {\footnotesize\textcolor{gray}{CVPR24}} 
                    & \ding{55} & \ding{55} & 60.40 & 35.18 & 58.24 & 71.20 & 42.71 & 20.09 & 6.72 \\
                \rowcolor[HTML]{FAFAFA}Layout Guidance \cite{chen2024training} 
                    & {\footnotesize\textcolor{gray}{WACV24}} 
                    & \ding{55} & \ding{51} & 40.01 & 38.80 & 95.95 & - & - & - & - \\
                \rowcolor[HTML]{FAFAFA}CONFORM\ding{168} \cite{meral2024conform} 
                    & {\footnotesize\textcolor{gray}{CVPR24}} 
                    & \ding{55} & \ding{55} & 60.73 & 38.48 & 62.33 & 73.01 & 45.82 & 25.57 & 9.52 \\
                \rowcolor[HTML]{FAFAFA}REVISION \cite{chatterjee2024revision} 
                    & {\footnotesize\textcolor{gray}{ECCV24}} 
                    & \ding{55} & \ding{51} & 53.96 & 52.71 & 97.69 & 77.79 & 61.02 & 44.90 & 27.15 \\
                \rowcolor[HTML]{FAFAFA}STORM \cite{han2025spatial} 
                    & {\footnotesize\textcolor{gray}{CVPR25}} 
                    & \ding{55} & \ding{55} & 61.01 & 57.58 & 94.39 & 85.93 & 69.71 & 49.01 & 25.70 \\
                \bottomrule

                \rowcolor[HTML]{FAFAFA}\textbf{InfSplign (Ours)} 
                    && \ding{55} & \ding{55} & \cellcolor[HTML]{E7F5F8}\textbf{67.36} & \cellcolor[HTML]{E7F5F8}\textbf{66.54} & \cellcolor[HTML]{E7F5F8}\textbf{98.79} & \cellcolor[HTML]{E7F5F8}\textbf{90.48} & \cellcolor[HTML]{E7F5F8}\textbf{77.79} & \cellcolor[HTML]{E7F5F8}\textbf{61.37} & \cellcolor[HTML]{E7F5F8}\textbf{36.54} \\
                \rowcolor[HTML]{FAFAFA}Improvement (Inference Time) 
                    &&&& \cellcolor[HTML]{E7F5F8}+6.35 & \cellcolor[HTML]{E7F5F8}+8.96 & \cellcolor[HTML]{E7F5F8}+1.10 & \cellcolor[HTML]{E7F5F8}+4.55 & \cellcolor[HTML]{E7F5F8}+8.08 & \cellcolor[HTML]{E7F5F8}+12.36 & \cellcolor[HTML]{E7F5F8}+9.39 \\
                \rowcolor[HTML]{FAFAFA}Improvement (All) 
                    &&&& \cellcolor[HTML]{E7F5F8}+1.80 & \cellcolor[HTML]{E7F5F8}+8.96 & \cellcolor[HTML]{E7F5F8}+1.10 & \cellcolor[HTML]{E7F5F8}+4.55 & \cellcolor[HTML]{E7F5F8}+8.08 & \cellcolor[HTML]{E7F5F8}+11.38 & \cellcolor[HTML]{E7F5F8}+7.63 \\
                \bottomrule
                \addlinespace[0.2ex]
                \bottomrule

                \rowcolor[HTML]{DCE7D9} \textbf{Stable Diffusion 2.1} &&&&&&&&&& \\
                \arrayrulecolor[HTML]{DCE7D9}
                \rowcolor[HTML]{FAFFF8} SD 2.1 \cite{rombach2022high} 
                    && - & \ding{55} & 47.83 & 30.25 & 63.24 & 64.42 & 35.74 & 16.13 & 4.70 \\
                \rowcolor[HTML]{FAFFF8}SPRIGHT \cite{chatterjee2024getting} 
                    & {\footnotesize\textcolor{gray}{ECCV24}} 
                    & \ding{51} & \ding{55} & 60.68 & 42.23 & 71.24 & 71.78 & 51.88 & 33.09 & 16.15 \\
                \rowcolor[HTML]{FAFFF8}REVISION \cite{chatterjee2024revision} 
                    & {\footnotesize\textcolor{gray}{ECCV24}} 
                    & \ding{55} & \ding{51} & 48.26 & 47.11 & 97.61 & 76.07 & 55.75 & 37.10 & 19.53 \\
                \rowcolor[HTML]{FAFFF8}STORM \cite{han2025spatial} 
                    & {\footnotesize\textcolor{gray}{CVPR25}} 
                    & \ding{55} & \ding{55} & 62.55 & 59.35 & 94.88 & 88.34 & 71.75 & 52.03 & 25.42 \\
                \rowcolor[HTML]{FAFFF8}CoMPaSS \cite{zhang2024compass} 
                    & {\footnotesize\textcolor{gray}{ICCV25}} 
                    & \ding{51} & \ding{55} & 68.22 & 62.06 & 90.96 & 85.02 & 71.29 & 56.03 & 35.90 \\
                \bottomrule

                \rowcolor[HTML]{FAFFF8}\textbf{InfSplign (Ours) }
                    && \ding{55} & \ding{55} & \cellcolor[HTML]{E7F5F8}\textbf{77.28} & \cellcolor[HTML]{E7F5F8}\textbf{76.26} & \cellcolor[HTML]{E7F5F8}\textbf{98.68} & \cellcolor[HTML]{E7F5F8}\textbf{94.65} & \cellcolor[HTML]{E7F5F8}\textbf{86.66} & \cellcolor[HTML]{E7F5F8}\textbf{73.48} & \cellcolor[HTML]{E7F5F8}\textbf{50.23} \\
                \rowcolor[HTML]{FAFFF8}Improvement (Inference Time) 
                    &&&& \cellcolor[HTML]{E7F5F8}+14.73 & \cellcolor[HTML]{E7F5F8}+16.91 & \cellcolor[HTML]{E7F5F8}+1.07 & \cellcolor[HTML]{E7F5F8}+6.31 & \cellcolor[HTML]{E7F5F8}+14.91 & \cellcolor[HTML]{E7F5F8}+21.45 & \cellcolor[HTML]{E7F5F8}+24.81 \\
                \rowcolor[HTML]{FAFFF8}Improvement (All) 
                    &&&&\cellcolor[HTML]{E7F5F8}+9.06 & \cellcolor[HTML]{E7F5F8}+14.20 & \cellcolor[HTML]{E7F5F8}+1.07 & \cellcolor[HTML]{E7F5F8}+6.31 & \cellcolor[HTML]{E7F5F8}+14.91 & \cellcolor[HTML]{E7F5F8}+17.45 & \cellcolor[HTML]{E7F5F8}+14.33 \\
                \bottomrule
            \end{tabular}
        }
    \end{threeparttable}
    \label{tab:visor_results}
\end{table*}

\begin{wraptable}{r}{0.5\textwidth}
    \centering
    \caption{Performance summary on T2I-CompBench on SD v$1.4$ and SD v$2.1$ backbone. \textbf{FT}: fine-tuning, \textbf{EI}: extra inputs. Best results are in \textbf{bold}.}
    \label{tab:t2i_results}
    \resizebox{\linewidth}{!}{
    \setlength{\tabcolsep}{2pt}
    \begin{tabular}{lccccc}
        \toprule
        \textbf{Method} &{Venue}& FT & EI & \cellcolor[HTML]{dedee0}\textbf{SD1.4} & \cellcolor[HTML]{DCE7D9}\textbf{SD2.1} \\
        \midrule
        Baseline \citeyearpar{rombach2022high} &&-&-&0.1246&0.1342\\
        Comp. Diff. \citeyearpar{liu2022compositional} &{\footnotesize\textcolor{gray}{ECCV22}}   &\ding{51}  &\ding{55}&\cellcolor[HTML]{FAFAFA}-&\cellcolor[HTML]{FAFFF8}0.0800\\
        Struc. Diff. \citeyearpar{feng2023trainingfree}&{\footnotesize\textcolor{gray}{ICLR23}}   &\ding{55}  &\ding{55}&\cellcolor[HTML]{FAFAFA}-&\cellcolor[HTML]{FAFFF8}0.1386\\
        Att.\&Ex. \citeyearpar{chefer2023attend}       &{\footnotesize\textcolor{gray}{SGRAPH23}} &\ding{55}  &\ding{55}&\cellcolor[HTML]{FAFAFA}-&\cellcolor[HTML]{FAFFF8}0.1455\\
        SPRIGHT \citeyearpar{chatterjee2024getting}    &{\footnotesize\textcolor{gray}{ECCV24}}   & \ding{51} &\ding{55}&\cellcolor[HTML]{FAFAFA}- &\cellcolor[HTML]{FAFFF8} 0.2133\\
        Revision \citeyearpar{chatterjee2024revision}  &{\footnotesize\textcolor{gray}{ECCV24}}   &\ding{55}  &\ding{51}&\cellcolor[HTML]{FAFAFA}0.3340&\cellcolor[HTML]{FAFFF8}-\\
        CoMPaSS \citeyearpar{zhang2024compass}         &{\footnotesize\textcolor{gray}{ICCV25}}   &\ding{51}  &\ding{55}&\cellcolor[HTML]{FAFAFA}0.3400 & \cellcolor[HTML]{FAFFF8}0.3200 \\
        STORM \citeyearpar{han2025spatial}             &{\footnotesize\textcolor{gray}{CVPR25}}   &\ding{55}  &\ding{55}&\cellcolor[HTML]{FAFAFA}0.1613 & \cellcolor[HTML]{FAFFF8}0.1981 \\
        \midrule
        InfSplign (Ours)&&\ding{55}&\ding{55}&\cellcolor[HTML]{E7F5F8}\bf{0.3771}	&\cellcolor[HTML]{E7F5F8}\bf{0.4172} \\
        \multicolumn{2}{l}{Improvement (Inference Time)}&&&\cellcolor[HTML]{E7F5F8}+4.31\%&\cellcolor[HTML]{E7F5F8}+21.91\%\\
        Improvement (All)&&&&\cellcolor[HTML]{E7F5F8}+3.71\%&\cellcolor[HTML]{E7F5F8}+9.72\%\\
        \bottomrule

    \end{tabular}
    }  
    \vspace{-\intextsep}
\end{wraptable}

\textbf{Quantitative Evaluation: T2I-CompBench.}
\autoref{tab:t2i_results} summarizes our performance compared to the state-of-the-art baselines both inference-time and fine-tuning on SD v$1.4$ and v$2.1$. Here again, \ourmethod{} consistently outperforms competitors by a margin. On SD~v$1.4$, it improves over the leading inference-time method REVISION by $+4.31\%$, while REVISION requires \emph{additional inputs}. Therefore, a more direct comparison is with STORM, in which \texttt{InfSplign} offers a significant $+21.58\%$ margin. Furthermore, our method even surpasses the strongest fine-tuning baseline in this setting, CoMPaSS, by $+3.71\%$. Following the same trend as in \autoref{tab:visor_results}, with the stronger backbone SD v$2.1$ our improvement margin increases significantly to $+21.91\%$ and $+9.72\%$ respectively.

\begin{wraptable}{r}{0.50\textwidth}
    \vspace{-\intextsep}
    \caption{The impact of \ourmethod{}'s loss terms $\mathcal{L}_{\text{spatial}}$ ($\mathcal{L}_{\text{s}}$), $\mathcal{L}_{\text{presence}}$ ($\mathcal{L}_{\text{p}}$) and $\mathcal{L}_{\text{balance}}$ ($\mathcal{L}_{\text{b}}$) on VISOR benchmark across two backbones. 
    } 
    \Large
    \centering
    \resizebox{\linewidth}{!}{
    \begin{tabular}{cccccccccc}
        \toprule
        \multirow{2}{*}{$\mathcal{L}_{\text{s}}$}&\multirow{2}{*}{$\mathcal{L}_{\text{p}}$}&\multirow{2}{*}{$\mathcal{L}_{\text{b}}$}&\multirow{2}{*}{OA (\%)}&\multicolumn{6}{c}{\textbf{VISOR (\%)}}\\
        \cmidrule(lr){5-10}
        &&&&uncond&cond&1&2&3&4\\
        \bottomrule
        
\rowcolor{gray!25}\multicolumn{5}{l}{\textbf{Stable Diffusion 1.4}}&\multicolumn{5}{l}{}\\

\rowcolor[HTML]{FAFAFA}\ding{55}&\ding{55}&\ding{55} & 32.74 & 20.38& 62.25  & 49.14 & 22.07 & 8.30 & 2.01 \\
\rowcolor[HTML]{FAFAFA}\ding{55}&\ding{55}&\ding{51} & 33.47 & 21.10& 63.03  & 50.34 & 23.26 & 8.79 & 1.99 \\
\rowcolor[HTML]{FAFAFA}\ding{55}&\ding{51}&\ding{55} & 53.54& 31.78 & 59.36  & 64.93 & 37.75 & 18.35 & 6.09 \\
\rowcolor[HTML]{FAFAFA}\ding{55}&\ding{51}&\ding{51} & 54.35 & 32.46& 59.73  & 65.93 & 38.72 & 18.84 & 6.37 \\
\rowcolor[HTML]{FAFAFA}\ding{51}&\ding{55}&\ding{55} & 59.33& 58.78 & 99.07  & 87.05 & 70.83 & 50.60 & 26.65 \\
\rowcolor[HTML]{FAFAFA}\ding{51}&\ding{55}&\ding{51} & 60.09 & 59.55 & 99.09 & 87.54 & 71.83 & 51.69 & 27.13 \\
\rowcolor[HTML]{FAFAFA}\ding{51}&\ding{51}&\ding{55} & 66.44& 65.63 & 98.78  & 89.69 & 77.18 & 60.01 & 35.62 \\
\midrule
\rowcolor[HTML]{E7F5F8}\ding{51}&\ding{51}&\ding{51} &\textbf{67.36} & \textbf{66.54}& \textbf{98.79 } & \textbf{90.48} & \textbf{77.79} & \textbf{61.37} & \textbf{36.54} \\
\midrule
\rowcolor[HTML]{DCE7D9}\multicolumn{5}{l}{\textbf{Stable Diffusion 2.1}}&\multicolumn{5}{l}{}\\

\rowcolor[HTML]{FAFFF8}\ding{55}&\ding{55}&\ding{55} & 44.37& 26.91 & 60.66  & 59.30 & 30.84 & 13.66 & 3.84 \\
\rowcolor[HTML]{FAFFF8}\ding{55}&\ding{55}&\ding{51} & 43.32 & 27.21 & 62.82 & 60.56 & 31.31 & 13.18 & 3.79 \\
\rowcolor[HTML]{FAFFF8}\ding{55}&\ding{51}&\ding{55} & 63.09& 37.17 & 58.91  & 71.65 & 44.98 & 23.38 & 8.67 \\
\rowcolor[HTML]{FAFFF8}\ding{55}&\ding{51}&\ding{51} & 62.85& 37.69 & 59.97  & 72.98 & 45.92 & 23.41 & 8.45 \\
\rowcolor[HTML]{FAFFF8}\ding{51}&\ding{55}&\ding{55} & 70.58& 69.70 & \textbf{98.76}  & 92.65 & 81.53 & 65.10 & 39.53 \\
\rowcolor[HTML]{FAFFF8}\ding{51}&\ding{55}&\ding{51} & 71.12& 70.18 & 98.69  & 92.90 & 82.20 & 65.76 & 39.86 \\
\rowcolor[HTML]{FAFFF8}\ding{51}&\ding{51}&\ding{55} & 76.20 & 75.21 & 98.71 & 94.12 & 85.81 & 72.33 & 48.59 \\
\midrule
\rowcolor[HTML]{E7F5F8}\ding{51}&\ding{51}&\ding{51} & \textbf{77.28}& \textbf{76.26}  & \textbf{98.68} &\textbf{ 94.65} &\textbf{ 86.66} & \textbf{73.48} & \textbf{50.23} \\
        \bottomrule
    \end{tabular}
    }
    \label{tab:losscomp}
    \vspace{-1\intextsep}
\end{wraptable}

\textbf{Ablation Studies.}
To better investigate the impact of the main (hyper)parameters and analyze the behavior of our loss functions, we conduct extensive studies. For this purpose, we use a subset of VISOR, which we refer to as $VISOR_{3160}$, in which instead of all $4$ spatial relations $R$ between two objects $A$ and $B$, we randomly sample a single relation. This results in $3160$ prompts that still cover all objects while keeping the dataset size manageable for ablations. Firstly, we perform a grid search over the key hyperparameters $\alpha$, $m$, $\lambda_{s}$, $\lambda_{p}$, and $\lambda_{b}$ for SD v$1.4$ and v$2.1$. The results of this extensive study are summarized in \autoref{tab:grid_search_sd1.4} and \autoref{tab:grid_search_sd2.1}, in \autoref{app:extend_ablations}. The optimal values obtained from this study are subsequently used for the experiments on the complete VISOR benchmark reported earlier in \autoref{tab:visor_results} and \autoref{tab:losscomp}, and the T2I-CompBench results in \autoref{tab:t2i_results}. 

\textbf{The effect of individual $\mathcal{L}_\texttt{InfSplign}$ terms.} \autoref{tab:losscomp} investigates the effect of the three loss terms across two backbones and demonstrates that all components contribute to improving baseline performance on both OA and VISOR metrics. $\mathcal{L}_{\text{spatial}}$ has the strongest impact on OA, and $\mathcal{L}_{\text{presence}}$ and $\mathcal{L}_{\text{balance}}$ come next in terms of impact in that order, incrementally improving the performance. As such, each and every component of the loss contributes to the overall strong margin beyond base SD v$1.4$ ($+34.62\%$). Combining $\mathcal{L}_{\text{spatial}}$ and $\mathcal{L}_{\text{presence}}$ seems to have a stronger impact than $\mathcal{L}_{\text{balance}}$ replacing the latter. This is expected as $\mathcal{L}_{\text{presence}}$ directly impacts OA, and thus the overall performance. Same trend applies to the other VISOR metrics as well as the stronger backbone SD v$2.1$.

\begin{figure}[t]
  \centering
  \includegraphics[width=0.95\linewidth]{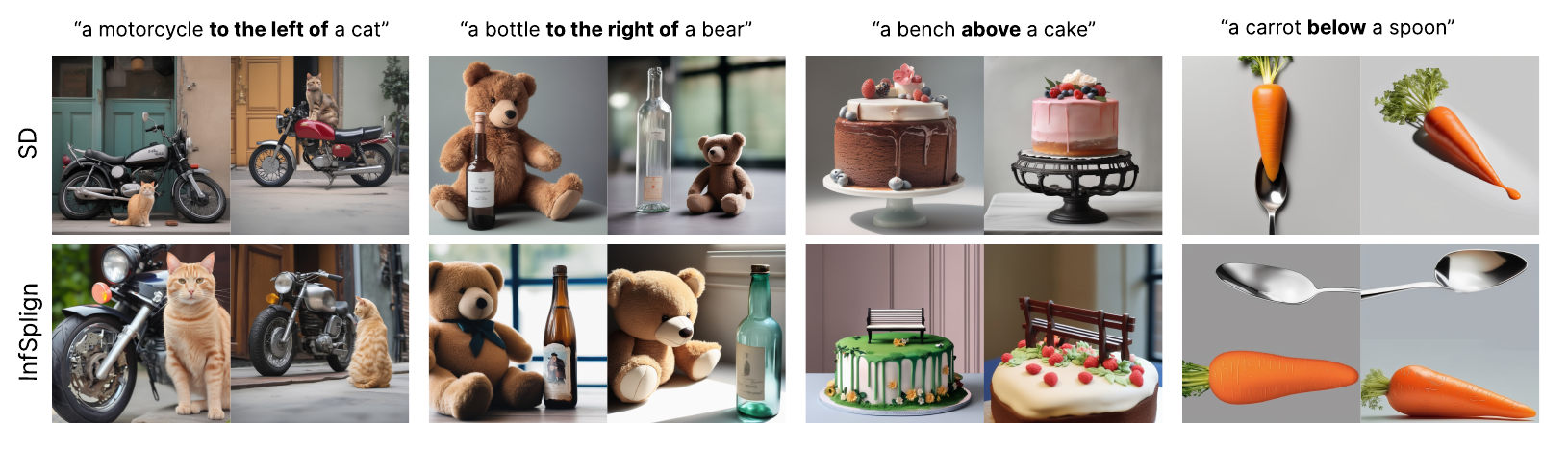}
  \caption{Qualitative comparison with SD across different VISOR prompts. }
  \label{fig:qualitative_success}
\end{figure}
\begin{figure}
  \centering
  \includegraphics[width=0.47\linewidth]{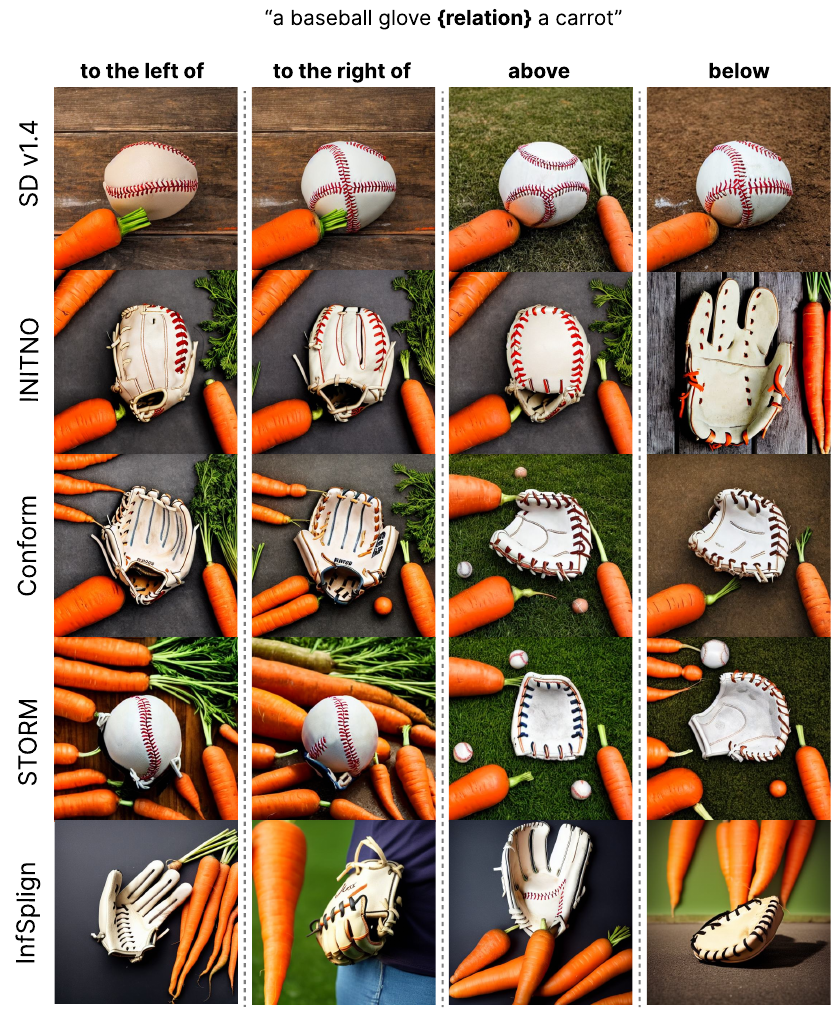}
  \includegraphics[width=0.47\linewidth]{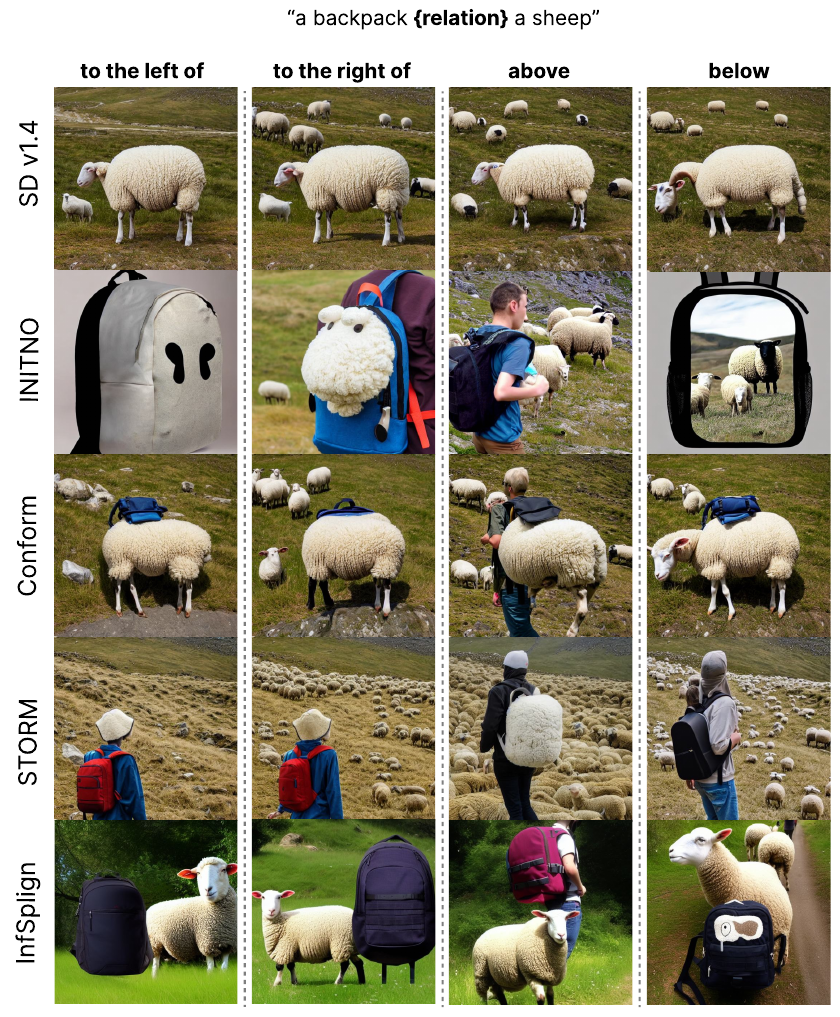}
  \caption{Comparison of spatial understanding across T2I diffusion models. \ourmethod{} consistently aligns objects according to the target relation better than SD v$1.4$ \cite{rombach2022high}, INITNO \cite{guo2024initno}, CONFORM \cite{meral2024conform} and STORM \cite{han2025spatial}.}
  \label{fig:model_comparison}
\end{figure}

\textbf{Qualitative Results.} \autoref{fig:qualitative_success} and \autoref{fig:model_comparison} shed light on the actual generation capabilities of \ourmethod{}. In \autoref{fig:qualitative_success} not only does the base SD fail to recognize the meaning of the spatial relationship from the prompt, but it also fails to generate both objects in unnatural object combinations. \ourmethod{} significantly improves upon these limitations and synthesizes spatially-aligned images even in atypical object settings \cite{park2024rare}. In \autoref{fig:model_comparison}, we illustrate the generated samples by the base SD and competing inference-time baselines INITNO, CONFORM, and STORM. The results across both prompts corroborate that \ourmethod{} honors spatial information the best among the competitors. Notably, the quality of generated images and subtle nuances therein can be further improved for all baselines with a stronger diffusion backbones such as SD v$2.1$ and SDXL. Additional examples  can be found in \autoref{app:extra_qual_results}.  

\setlength{\textfloatsep}{10pt} 
\setlength{\floatsep}{5pt}     
\setlength{\intextsep}{5pt}    

\section{Concluding Remarks}
In this work, we tackle the challenge of spatial understanding in T2I diffusion models, a persistent limitation that undermines accurate compositional generation. We introduce \texttt{InfSplign}, a lightweight yet powerful inference-time method that improves spatial alignment and object preservation without requiring retraining or external supervision. The proposed approach applies a spatial loss during sampling, guided by attention maps to estimate object locations and variances. This not only enforces spatial relations specified in the prompt, but also preserves object fidelity by regulating attention map variances. Crucially, \texttt{InfSplign} operates without modifying model weights or depending on auxiliary inputs such as bounding boxes or segmentation masks, making it fully modular and applicable to any diffusion model. Despite its simplicity, it surpasses fine-tuning-based methods and prior inference-time strategies aimed at spatial consistency, demonstrating that reliable object placement can be achieved without model retraining. 


\textbf{Limitations.} The base diffusion model often struggles to generate uncommon object combinations posing a challenge to any inference-time approach. For instance, VISOR includes a wide range of object pairs, many of which rarely co-occur in natural scenes. While this setup is valuable for testing generalization in spatial reasoning, it also makes the task more difficult, particularly when one or both objects fail to appear in the generated image. In such cases, \texttt{InfSplign} (and other inference-time baselines) cannot always effectively correct the spatial alignment, as the prerequisite object presence is impaired. As a result, object accuracy becomes the bottleneck in such scenarios.

\newpage
\bibliography{references}
\bibliographystyle{apalike}

\newpage
\appendix
\newpage

\section{Benchmarks and metrics}\label{app:benchmarks}

\subsection{VISOR}
VISOR \cite{visor} uses 80 object categories from the MS-COCO data set and defines unordered pairs of all objects, resulting in 6320 object combinations. Each object pair is matching with all 4 spatial relationships - "to the left of", "to the right of", "above" and "below". This gives us a total of 25280 spatial prompts. The VISOR dataset includes 6320 more prompts for each of the object pairs and the concept conjunction "and". Finally, it includes 80 more one-object prompts for all 80 COCO categories. In total, this results in 31680 text prompts. The VISOR benchmark defines the VISOR 1-4 scores, hence 4 images are generated per prompt. All VISOR images are generated with seed 42, ensuring that the images across prompts start from the same initial noise. We followed these guidelines in the evaluation of \ourmethod{}. 

The VISOR benchmark uses the OWL-ViT object detector to localize objects classes which are matched to the objects from the prompt. Then using a simple set of rules it predicts the spatial relationship and compares it with the ground truth relationship to determine if the image aligns with the spatial information in the prompt. The benchmark defines the following set of metrics: OA (object accuracy) indicating the number of images in which both objects were generated and detected, $\text{VISOR}_\text{cond}$ estimates the conditional probability of how many of the images adhere to the spatial relationship given that both objects are generated correctly, VISOR (or $\text{VISOR}_\text{uncond}$) gives a ratio of the number of images generated with a correct spatial relationship independent of the two objects being generated correctly. The $\text{VISOR}_1$ to $\text{VISOR}_N$ scores define the percentage of images out of N that have the correct spatial relationship between the objects present. All VISOR metrics are reported as a percentage.

\subsection{T2I-CompBench}
T2I-CompBench \cite{huang2023t2i} is a for compositional T2I generation benchmark. It looks at attribute binding, object relationships and complex compositions. For our research, we focus on spatial alignment, hence focus on the 2D spatial relationships subcategory. The benchmark offers $1000$ spatial prompts, but only $300$ of them are used for testing since the authors present a fine-tuning approach and use the remaining $700$ prompts for training. T2I-CompBench includes 7 spatial relationships. The first 4 relationships are directional and consistent with VISOR, while the last 3 are relative, requiring close proximity of objects: \textit{"on the left of"}, \textit{"on the right of"}, \textit{"on the top of"},\textit{ "on the bottom of"}, \textit{"on the side of"}, \textit{"next to"}, \textit{"near"}. The objects are chosen from 3 categories: persons, animals and objects. In the benchmark the $10$ images per prompt are generated with 10 different random seeds: $42 ... 51$, again ensuring that across prompts, the images start with the same reproducible initial noise. 

T2I-CompBench uses the UniDet object detector to predict the location of each generated object.
The centers of the objects are calculated from the detected bounding boxes. The metric estimates whether the spatial relationship is respected between both objects by comparing the centroids, measuring in which axis the objects are further apart, and the intersection-over-union (IoU), to avoid object overlap. 
To address the correctness of the 3 relative spatial relationships, the metric is calculated a weighted combination of confidence score for each of the two detected objects and the positional score.

\newpage
\section{Extended Ablation Studies}
\label{app:extend_ablations}
To understand the effect of the interplay of hyperparameters on the spatial alignment score, a gridsearch for the parameters \textbf{$\alpha$}, \textbf{$m$}, \textbf{$\lambda_{s}$}, \textbf{$\lambda_{p}$}, \textbf{$\lambda_{b}$} has been conducted. All experiments have been conducted on the $VISOR_{3160}$ subset, to keep experiments computationally feasible. The selected hyperparameters for SD v$1.4$ (\autoref{tab:grid_search_sd1.4}) are: $\alpha = 1.5$, $m = 0.25$, $\lambda_{\text{s}}{=}0.5$, $\lambda_{\text{p}}{=}1$, $\lambda_{\text{b}}{=}0.5$. Alternatively the selected hyperparameters for SD v$2.1$ (\autoref{tab:grid_search_sd2.1}) are: $\alpha = 1.5$, $m = 0.5$, $\lambda_{\text{s}}{=}0.5$, $\lambda_{\text{p}}{=}1$, $\lambda_{\text{b}}{=}1.0$.

\begin{table}[h!]
\centering
\caption{Grid search results for hyperparameter optimization for SD v$1.4$ on $VISOR_{3160}$}
\label{tab:grid_search_sd1.4}
\resizebox{0.8\textwidth}{!}{%
\begin{tabular}{|c|c|c|c|c|c|c|c|c|c|c|c|}
\hline
\rowcolor{gray!25}
\textbf{$\alpha$} & \textbf{$m$} & \textbf{$\lambda_{s}$} & \textbf{$\lambda_{p}$} & \textbf{$\lambda_{b}$} & \textbf{OA} & \textbf{VISOR$_{uncond}$} & \textbf{VISOR$_{cond}$} & \textbf{VISOR$_1$} & \textbf{VISOR$_2$} & \textbf{VISOR$_3$} & \textbf{VISOR$_4$} \\
\hline
\rowcolor[HTML]{FAFAFA}
0.5 & 0.50 & 0.5 & 0.5 & 0.5 & 58.68 & 55.27 & 94.19 & 83.23 & 66.36 & 46.46 & 25.03 \\
\rowcolor[HTML]{FAFAFA}
0.5 & 0.50 & 0.5 & 0.5 & 1.0 & 59.04 & 55.97 & 94.80 & 84.21 & 66.74 & 47.75 & 25.16 \\
\rowcolor[HTML]{FAFAFA}
0.5 & 0.50 & 0.5 & 1.0 & 0.5 & 59.03 & 53.75 & 91.06 & 82.06 & 64.72 & 45.60 & 22.63 \\
\rowcolor[HTML]{FAFAFA}
0.5 & 0.50 & 0.5 & 1.0 & 1.0 & 59.98 & 54.88 & 91.51 & 83.86 & 65.73 & 46.11 & 23.83 \\
\rowcolor[HTML]{FAFAFA}
0.5 & 0.50 & 1.0 & 0.5 & 0.5 & 64.72 & 63.65 & 98.34 & 88.61 & 75.48 & 56.93 & 33.58 \\
\rowcolor[HTML]{FAFAFA}
0.5 & 0.50 & 1.0 & 0.5 & 1.0 & 65.08 & 64.00 & 98.35 & 89.37 & 75.44 & 58.07 & 33.13 \\
\rowcolor[HTML]{FAFAFA}
0.5 & 0.50 & 1.0 & 1.0 & 0.5 & 64.42 & 62.96 & 97.73 & 87.22 & 73.89 & 57.63 & 33.10 \\
\rowcolor[HTML]{FAFAFA}
0.5 & 0.50 & 1.0 & 1.0 & 1.0 & 65.12 & 63.58 & 97.63 & 87.63 & 74.56 & 58.29 & 33.83 \\
\hline
\rowcolor[HTML]{FAFAFA}
0.5 & 0.75 & 0.5 & 0.5 & 0.5 & 59.50 & 56.66 & 95.23 & 84.46 & 67.60 & 48.73 & 25.85 \\
\rowcolor[HTML]{FAFAFA}
0.5 & 0.75 & 0.5 & 0.5 & 1.0 & 60.32 & 57.62 & 95.53 & 85.29 & 69.27 & 49.08 & 26.84 \\
\rowcolor[HTML]{FAFAFA}
0.5 & 0.75 & 0.5 & 1.0 & 0.5 & 59.88 & 55.57 & 92.80 & 83.29 & 66.23 & 47.88 & 24.87 \\
\rowcolor[HTML]{FAFAFA}
0.5 & 0.75 & 0.5 & 1.0 & 1.0 & 60.83 & 56.80 & 93.38 & 84.40 & 68.13 & 48.64 & 26.04 \\
\rowcolor[HTML]{FAFAFA}
0.5 & 0.75 & 1.0 & 0.5 & 0.5 & 65.23 & 64.45 & 98.81 & 88.96 & 76.80 & 58.73 & 33.32 \\
\rowcolor[HTML]{FAFAFA}
0.5 & 0.75 & 1.0 & 0.5 & 1.0 & 65.92 & 65.19 & 98.90 & 89.62 & 77.37 & 59.46 & 34.30 \\
\rowcolor[HTML]{FAFAFA}
0.5 & 0.75 & 1.0 & 1.0 & 0.5 & 65.77 & 64.58 & 98.20 & 88.96 & 75.92 & 58.80 & 34.65 \\
\rowcolor[HTML]{FAFAFA}
0.5 & 0.75 & 1.0 & 1.0 & 1.0 & 65.32 & 64.12 & 98.16 & 88.32 & 75.35 & 58.32 & 34.49 \\
\hline
\rowcolor[HTML]{FAFAFA}
1.0 & 0.25 & 0.5 & 0.5 & 0.5 & 64.28 & 63.20 & 98.31 & 87.91 & 74.72 & 57.15 & 33.01 \\
\rowcolor[HTML]{FAFAFA}
1.0 & 0.25 & 0.5 & 0.5 & 1.0 & 64.57 & 63.42 & 98.22 & 88.73 & 74.97 & 57.31 & 32.66 \\
\rowcolor[HTML]{FAFAFA}
1.0 & 0.25 & 0.5 & 1.0 & 0.5 & 64.72 & 63.24 & 97.70 & 87.91 & 74.68 & 57.06 & 33.29 \\
\rowcolor[HTML]{FAFAFA}
1.0 & 0.25 & 0.5 & 1.0 & 1.0 & 65.02 & 63.51 & 97.68 & 88.04 & 74.18 & 57.72 & 34.11 \\
\rowcolor[HTML]{FAFAFA}
1.0 & 0.25 & 1.0 & 0.5 & 0.5 & 65.97 & 65.44 & 99.21 & 90.73 & 78.01 & 59.81 & 33.23 \\
\rowcolor[HTML]{FAFAFA}
1.0 & 0.25 & 1.0 & 0.5 & 1.0 & 66.46 & 65.94 & 99.23 & 90.98 & 78.13 & 60.85 & 33.80 \\
\rowcolor[HTML]{FAFAFA}
1.0 & 0.25 & 1.0 & 1.0 & 0.5 & 67.74 & 67.24 & 99.26 & 90.63 & 78.96 & 62.18 & 37.18 \\
\rowcolor[HTML]{FAFAFA}
1.0 & 0.25 & 1.0 & 1.0 & 1.0 & 67.90 & \underline{67.40} & 99.25 & 91.33 & 79.30 & 62.09 & 36.87 \\
\hline
\rowcolor[HTML]{FAFAFA}
1.0 & 0.50 & 0.5 & 0.5 & 0.5 & 65.64 & 64.87 & 98.82 & 89.65 & 76.52 & 59.24 & 34.05 \\
\rowcolor[HTML]{FAFAFA}
1.0 & 0.50 & 0.5 & 0.5 & 1.0 & 66.38 & 65.62 & 98.86 & 90.25 & 77.69 & 59.65 & 34.87 \\
\rowcolor[HTML]{FAFAFA}
1.0 & 0.50 & 0.5 & 1.0 & 0.5 & 66.35 & 65.30 & 98.41 & 89.15 & 76.71 & 59.78 & 35.57 \\
\rowcolor[HTML]{FAFAFA}
1.0 & 0.50 & 0.5 & 1.0 & 1.0 & 66.37 & 65.25 & 98.32 & 89.75 & 76.01 & 59.65 & 35.60 \\
\rowcolor[HTML]{FAFAFA}
1.0 & 0.50 & 1.0 & 0.5 & 0.5 & 63.67 & 63.26 & 99.35 & 90.03 & 75.76 & 56.39 & 30.85 \\
\rowcolor[HTML]{FAFAFA}
1.0 & 0.50 & 1.0 & 0.5 & 1.0 & 63.70 & 63.20 & 99.22 & 90.19 & 75.70 & 56.49 & 30.41 \\
\rowcolor[HTML]{FAFAFA}
1.0 & 0.50 & 1.0 & 1.0 & 0.5 & 65.88 & 65.37 & 99.23 & 91.01 & 77.88 & 59.18 & 33.42 \\
\rowcolor[HTML]{FAFAFA}
1.0 & 0.50 & 1.0 & 1.0 & 1.0 & 66.60 & 66.17 & 99.36 & 91.39 & 78.83 & 59.75 & 34.72 \\
\hline
\rowcolor[HTML]{FAFAFA}
1.0 & 0.75 & 0.5 & 0.5 & 0.5 & 66.18 & 65.51 & 99.00 & 90.29 & 77.79 & 60.22 & 33.77 \\
\rowcolor[HTML]{FAFAFA}
1.0 & 0.75 & 0.5 & 0.5 & 1.0 & 66.14 & 65.51 & 99.04 & 90.85 & 78.58 & 60.06 & 32.53 \\
\rowcolor[HTML]{FAFAFA}
1.0 & 0.75 & 0.5 & 1.0 & 0.5 & 66.44 & 65.56 & 98.68 & 90.03 & 77.15 & 60.67 & 34.40 \\
\rowcolor[HTML]{FAFAFA}
1.0 & 0.75 & 0.5 & 1.0 & 1.0 & 67.62 & 66.76 & 98.73 & 90.29 & 78.26 & 61.99 & 36.49 \\
\rowcolor[HTML]{FAFAFA}
1.0 & 0.75 & 1.0 & 0.5 & 0.5 & 61.05 & 60.64 & 99.33 & 88.29 & 73.29 & 53.01 & 27.98 \\
\rowcolor[HTML]{FAFAFA}
1.0 & 0.75 & 1.0 & 0.5 & 1.0 & 61.78 & 61.41 & 99.40 & 90.10 & 74.43 & 53.10 & 28.01 \\
\rowcolor[HTML]{FAFAFA}
1.0 & 0.75 & 1.0 & 1.0 & 0.5 & 63.39 & 63.02 & 99.43 & 90.41 & 75.67 & 55.85 & 30.16 \\
\rowcolor[HTML]{FAFAFA}
1.0 & 0.75 & 1.0 & 1.0 & 1.0 & 63.70 & 63.30 & 99.37 & 90.63 & 76.20 & 55.98 & 30.38 \\
\hline
\rowcolor[HTML]{FAFAFA}
1.5 & 0.25 & 0.5 & 0.5 & 0.5 & 66.46 & 65.95 & 99.23 & 90.06 & 78.17 & 60.70 & 34.87 \\
\rowcolor[HTML]{FAFAFA}
1.5 & 0.25 & 0.5 & 0.5 & 1.0 & 66.70 & 66.08 & 99.06 & 90.41 & 79.30 & 60.38 & 34.21 \\
\rowcolor[HTML]{E7F5F8}
1.5 & 0.25 & 0.5 & 1.0 & 0.5 & \underline{67.96} & 67.26 & 98.96 & 90.38 & 78.54 & \underline{62.79} & 37.31 \\
\rowcolor[HTML]{FAFAFA}
1.5 & 0.25 & 0.5 & 1.0 & 1.0 & 67.94 & 67.20 & 98.92 & 90.41 & 78.92 & 61.77 & \underline{37.69} \\
\rowcolor[HTML]{FAFAFA}
1.5 & 0.25 & 1.0 & 0.5 & 0.5 & 61.61 & 61.16 & 99.26 & 89.49 & 73.54 & 53.51 & 28.07 \\
\rowcolor[HTML]{FAFAFA}
1.5 & 0.25 & 1.0 & 0.5 & 1.0 & 62.38 & 61.96 & 99.33 & 89.40 & 75.76 & 54.53 & 28.17 \\
\rowcolor[HTML]{FAFAFA}
1.5 & 0.25 & 1.0 & 1.0 & 0.5 & 64.11 & 63.73 & 99.42 & 90.29 & 76.71 & 57.34 & 30.60 \\
\rowcolor[HTML]{FAFAFA}
1.5 & 0.25 & 1.0 & 1.0 & 1.0 & 64.80 & 64.35 & 99.30 & 90.89 & 77.44 & 57.56 & 31.52 \\
\hline
\rowcolor[HTML]{FAFAFA}
1.5 & 0.50 & 0.5 & 0.5 & 0.5 & 64.98 & 64.53 & 99.32 & 90.19 & 77.15 & 58.39 & 32.41 \\
\rowcolor[HTML]{FAFAFA}
1.5 & 0.50 & 0.5 & 0.5 & 1.0 & 65.55 & 65.08 & 99.29 & 90.98 & 77.72 & 59.30 & 32.31 \\
\rowcolor[HTML]{FAFAFA}
1.5 & 0.50 & 0.5 & 1.0 & 0.5 & 67.56 & 66.92 & 99.05 & 91.30 & 78.89 & 60.92 & 36.55 \\
\rowcolor[HTML]{FAFAFA}
1.5 & 0.50 & 0.5 & 1.0 & 1.0 & 67.56 & 67.08 & 99.30 & \underline{91.49} & \underline{79.56} & 61.77 & 35.51 \\
\rowcolor[HTML]{FAFAFA}
1.5 & 0.50 & 1.0 & 0.5 & 0.5 & 53.77 & 53.46 & 99.43 & 84.78 & 65.29 & 43.29 & 20.48 \\
\rowcolor[HTML]{FAFAFA}
1.5 & 0.50 & 1.0 & 0.5 & 1.0 & 54.08 & 53.72 & 99.33 & 85.32 & 66.14 & 43.54 & 19.87 \\
\rowcolor[HTML]{FAFAFA}
1.5 & 0.50 & 1.0 & 1.0 & 0.5 & 56.95 & 56.65 & 99.47 & 87.44 & 68.86 & 47.28 & 23.01 \\
\rowcolor[HTML]{FAFAFA}
1.5 & 0.50 & 1.0 & 1.0 & 1.0 & 57.29 & 56.95 & 99.42 & 87.47 & 69.56 & 46.90 & 23.89 \\
\hline
\end{tabular}%
}
\end{table}

\begin{table}[p]
\centering
\caption{Grid search results for hyperparameter optimization for SDv$2.1$ on $VISOR_{3160}$}
\label{tab:grid_search_sd2.1}
\resizebox{0.8\textwidth}{!}{%
\begin{tabular}{|c|c|c|c|c|c|c|c|c|c|c|c|}
\hline
\rowcolor[HTML]{DCE7D9}
\textbf{$\alpha$} & \textbf{$m$} & \textbf{$\lambda_{s}$} ($\mathcal{L}_{\text{spatial}}$) & \textbf{$\lambda_{p}$} ($\mathcal{L}_{\text{presence}}$) & \textbf{$\lambda_{b}$} (($\mathcal{L}_{\text{balance}}$)) & \textbf{OA} & \textbf{VISOR$_{uncond}$} & \textbf{VISOR$_{cond}$} & \textbf{VISOR$_1$} & \textbf{VISOR$_2$} & \textbf{VISOR$_3$} & \textbf{VISOR$_4$} \\
\hline
\rowcolor[HTML]{FAFFF8}
0.5 & 0.50 & 0.5 & 0.5 & 0.5 & 66.61 & 62.99 & 94.56 & 87.75 & 73.45 & 56.68 & 34.08 \\
\rowcolor[HTML]{FAFFF8}
0.5 & 0.50 & 0.5 & 0.5 & 1.0 & 66.52 & 63.16 & 94.95 & 88.70 & 74.43 & 56.49 & 33.01 \\
\rowcolor[HTML]{FAFFF8}
0.5 & 0.50 & 0.5 & 1.0 & 0.5 & 68.88 & 63.52 & 92.21 & 88.83 & 74.11 & 56.52 & 34.62 \\
\rowcolor[HTML]{FAFFF8}
0.5 & 0.50 & 0.5 & 1.0 & 1.0 & 69.24 & 64.45 & 93.08 & 89.43 & 75.41 & 57.85 & 35.10 \\
\rowcolor[HTML]{FAFFF8}
0.5 & 0.50 & 1.0 & 0.5 & 0.5 & 72.63 & 71.15 & 97.96 & 92.37 & 81.61 & 66.77 & 43.83 \\
\rowcolor[HTML]{FAFFF8}
0.5 & 0.50 & 1.0 & 0.5 & 1.0 & 71.83 & 70.36 & 97.95 & 92.25 & 80.98 & 65.19 & 43.01 \\
\rowcolor[HTML]{FAFFF8}
0.5 & 0.50 & 1.0 & 1.0 & 0.5 & 74.55 & 72.63 & 97.42 & 92.91 & 82.63 & 68.29 & 46.68 \\
\rowcolor[HTML]{FAFFF8}
0.5 & 0.50 & 1.0 & 1.0 & 1.0 & 75.11 & 73.26 & 97.54 & 93.73 & 82.98 & 69.24 & 47.09 \\
\hline
\rowcolor[HTML]{FAFFF8}
0.5 & 0.75 & 0.5 & 0.5 & 0.5 & 67.81 & 64.92 & 95.74 & 89.37 & 75.25 & 58.73 & 36.33 \\
\rowcolor[HTML]{FAFFF8}
0.5 & 0.75 & 0.5 & 0.5 & 1.0 & 68.23 & 65.21 & 95.57 & 89.56 & 75.76 & 58.99 & 36.52 \\
\rowcolor[HTML]{FAFFF8}
0.5 & 0.75 & 0.5 & 1.0 & 0.5 & 70.20 & 65.89 & 93.87 & 89.40 & 76.84 & 59.97 & 37.37 \\
\rowcolor[HTML]{FAFFF8}
0.5 & 0.75 & 0.5 & 1.0 & 1.0 & 70.43 & 66.55 & 94.50 & 90.63 & 77.44 & 60.63 & 37.50 \\
\rowcolor[HTML]{FAFFF8}
0.5 & 0.75 & 1.0 & 0.5 & 0.5 & 73.88 & 72.51 & 98.15 & 92.53 & 82.82 & 68.48 & 46.20 \\
\rowcolor[HTML]{FAFFF8}
0.5 & 0.75 & 1.0 & 0.5 & 1.0 & 73.62 & 72.29 & 98.20 & 93.20 & 82.72 & 68.29 & 44.97 \\
\rowcolor[HTML]{FAFFF8}
0.5 & 0.75 & 1.0 & 1.0 & 0.5 & 75.65 & 74.18 & 98.06 & 93.13 & 83.96 & 70.98 & 48.64 \\
\rowcolor[HTML]{FAFFF8}
0.5 & 0.75 & 1.0 & 1.0 & 1.0 & 74.92 & 73.36 & 97.92 & 93.17 & 83.26 & 69.27 & 47.75 \\
\hline
\rowcolor[HTML]{FAFFF8}
1.0 & 0.25 & 0.5 & 0.5 & 0.5 & 72.04 & 70.63 & 98.05 & 92.06 & 81.08 & 66.36 & 43.04 \\
\rowcolor[HTML]{FAFFF8}
1.0 & 0.25 & 0.5 & 0.5 & 1.0 & 72.10 & 70.74 & 98.12 & 91.96 & 82.03 & 66.42 & 42.56 \\
\rowcolor[HTML]{FAFFF8}
1.0 & 0.25 & 0.5 & 1.0 & 0.5 & 74.32 & 72.41 & 97.44 & 93.04 & 82.18 & 68.45 & 45.98 \\
\rowcolor[HTML]{FAFFF8}
1.0 & 0.25 & 0.5 & 1.0 & 1.0 & 74.79 & 72.94 & 97.51 & 93.42 & 82.85 & 68.70 & 46.77 \\
\rowcolor[HTML]{FAFFF8}
1.0 & 0.25 & 1.0 & 0.5 & 0.5 & 75.40 & 74.51 & 98.83 & 94.08 & 85.03 & 71.27 & 47.66 \\
\rowcolor[HTML]{FAFFF8}
1.0 & 0.25 & 1.0 & 0.5 & 1.0 & 75.22 & 74.29 & 98.76 & 93.99 & 85.70 & 71.11 & 46.36 \\
\rowcolor[HTML]{FAFFF8}
1.0 & 0.25 & 1.0 & 1.0 & 0.5 & 76.98 & 76.02 & 98.76 & 94.68 & 85.63 & 73.20 & 50.57 \\
\rowcolor[HTML]{FAFFF8}
1.0 & 0.25 & 1.0 & 1.0 & 1.0 & 77.89 & 76.95 & 98.80 & 95.16 & 87.18 & \underline{75.35} & 50.13 \\
\hline
\rowcolor[HTML]{FAFFF8}
1.0 & 0.50 & 0.5 & 0.5 & 0.5 & 74.28 & 72.97 & 98.23 & 93.07 & 83.29 & 69.91 & 45.60 \\
\rowcolor[HTML]{FAFFF8}
1.0 & 0.50 & 0.5 & 0.5 & 1.0 & 73.97 & 72.81 & 98.43 & 93.70 & 83.54 & 68.67 & 45.32 \\
\rowcolor[HTML]{FAFFF8}
1.0 & 0.50 & 0.5 & 1.0 & 0.5 & 75.69 & 74.23 & 98.08 & 93.58 & 83.83 & 71.04 & 48.48 \\
\rowcolor[HTML]{FAFFF8}
1.0 & 0.50 & 0.5 & 1.0 & 1.0 & 76.16 & 74.70 & 98.09 & 93.96 & 85.35 & 70.38 & 49.11 \\
\rowcolor[HTML]{FAFFF8}
1.0 & 0.50 & 1.0 & 0.5 & 0.5 & 74.34 & 73.68 & 99.11 & 94.11 & 84.94 & 70.51 & 45.16 \\
\rowcolor[HTML]{FAFFF8}
1.0 & 0.50 & 1.0 & 0.5 & 1.0 & 74.67 & 73.90 & 98.97 & 93.99 & 85.10 & 70.76 & 45.76 \\
\rowcolor[HTML]{FAFFF8}
1.0 & 0.50 & 1.0 & 1.0 & 0.5 & 76.89 & 75.97 & 98.81 & 95.06 & 86.77 & 73.54 & 48.51 \\
\rowcolor[HTML]{FAFFF8}
1.0 & 0.50 & 1.0 & 1.0 & 1.0 & 76.69 & 75.84 & 98.90 & \underline{95.22} & 86.61 & 73.17 & 48.35 \\
\hline
\rowcolor[HTML]{FAFFF8}
1.0 & 0.75 & 0.5 & 0.5 & 0.5 & 74.59 & 73.53 & 98.58 & 93.64 & 84.02 & 69.84 & 46.61 \\
\rowcolor[HTML]{FAFFF8}
1.0 & 0.75 & 0.5 & 0.5 & 1.0 & 75.09 & 73.99 & 98.54 & 93.99 & 84.72 & 70.95 & 46.30 \\
\rowcolor[HTML]{FAFFF8}
1.0 & 0.75 & 0.5 & 1.0 & 0.5 & 76.62 & 75.37 & 98.37 & 93.61 & 85.63 & 72.94 & 49.30 \\
\rowcolor[HTML]{FAFFF8}
1.0 & 0.75 & 0.5 & 1.0 & 1.0 & 77.15 & 76.04 & 98.56 & 94.62 & 86.61 & 73.10 & 49.84 \\
\rowcolor[HTML]{FAFFF8}
1.0 & 0.75 & 1.0 & 0.5 & 0.5 & 72.90 & 72.15 & 98.97 & 93.96 & 84.27 & 68.13 & 42.25 \\
\rowcolor[HTML]{FAFFF8}
1.0 & 0.75 & 1.0 & 0.5 & 1.0 & 73.10 & 72.40 & 99.04 & 94.24 & 84.18 & 69.43 & 41.74 \\
\rowcolor[HTML]{FAFFF8}
1.0 & 0.75 & 1.0 & 1.0 & 0.5 & 74.98 & 74.11 & 98.85 & 93.96 & 86.08 & 71.17 & 45.25 \\
\rowcolor[HTML]{FAFFF8}
1.0 & 0.75 & 1.0 & 1.0 & 1.0 & 75.25 & 74.42 & 98.89 & 94.84 & 85.51 & 71.33 & 45.98 \\
\hline
\rowcolor[HTML]{FAFFF8}
1.5 & 0.25 & 0.5 & 0.5 & 0.5 & 75.06 & 74.13 & 98.76 & 93.48 & 84.62 & 71.11 & 47.31 \\
\rowcolor[HTML]{FAFFF8}
1.5 & 0.25 & 0.5 & 0.5 & 1.0 & 75.11 & 74.11 & 98.66 & 94.05 & 85.41 & 70.57 & 46.39 \\
\rowcolor[HTML]{FAFFF8}
1.5 & 0.25 & 0.5 & 1.0 & 0.5 & 76.69 & 75.49 & 98.44 & 93.86 & 85.57 & 72.75 & 49.78 \\
\rowcolor[HTML]{FAFFF8}
1.5 & 0.25 & 0.5 & 1.0 & 1.0 & 76.84 & 75.77 & 98.61 & 93.83 & 85.51 & 72.94 & 50.79 \\
\rowcolor[HTML]{FAFFF8}
1.5 & 0.25 & 1.0 & 0.5 & 0.5 & 72.24 & 71.62 & \underline{99.15} & 94.02 & 83.70 & 67.53 & 41.23 \\
\rowcolor[HTML]{FAFFF8}
1.5 & 0.25 & 1.0 & 0.5 & 1.0 & 72.22 & 71.46 & 98.95 & 93.35 & 83.67 & 67.85 & 40.95 \\
\rowcolor[HTML]{FAFFF8}
1.5 & 0.25 & 1.0 & 1.0 & 0.5 & 74.78 & 74.05 & 99.03 & 94.91 & 85.54 & 71.17 & 44.59 \\
\rowcolor[HTML]{FAFFF8}
1.5 & 0.25 & 1.0 & 1.0 & 1.0 & 74.80 & 73.99 & 98.91 & 94.49 & 85.73 & 70.89 & 44.84 \\
\hline
\rowcolor[HTML]{FAFFF8}
1.5 & 0.50 & 0.5 & 0.5 & 0.5 & 75.01 & 74.12 & 98.82 & 93.89 & 84.97 & 70.98 & 46.65 \\
\rowcolor[HTML]{FAFFF8}
1.5 & 0.50 & 0.5 & 0.5 & 1.0 & 75.06 & 74.19 & 98.83 & 94.08 & 85.19 & 71.04 & 46.42 \\
\rowcolor[HTML]{FAFFF8}
1.5 & 0.50 & 0.5 & 1.0 & 0.5 & 77.36 & 76.43 & 98.80 & 94.62 & 86.36 & 73.89 & 50.85 \\
\rowcolor[HTML]{E7F5F8}
1.5 & 0.50 & 0.5 & 1.0 & 1.0 & \underline{77.98} & \underline{77.10} & 98.86 & 95.00 & \underline{87.18} & 74.40 & \underline{51.80} \\
\rowcolor[HTML]{FAFFF8}
1.5 & 0.50 & 1.0 & 0.5 & 0.5 & 67.41 & 66.77 & 99.05 & 92.18 & 80.00 & 61.74 & 33.17 \\
\rowcolor[HTML]{FAFFF8}
1.5 & 0.50 & 1.0 & 0.5 & 1.0 & 66.80 & 66.11 & 98.97 & 91.42 & 79.34 & 60.35 & 33.32 \\
\rowcolor[HTML]{FAFFF8}
1.5 & 0.50 & 1.0 & 1.0 & 0.5 & 69.17 & 68.55 & 99.11 & 93.42 & 81.23 & 63.39 & 36.17 \\
\rowcolor[HTML]{FAFFF8}
1.5 & 0.50 & 1.0 & 1.0 & 1.0 & 69.22 & 68.58 & 99.09 & 93.17 & 81.11 & 63.45 & 36.61 \\
\hline
\end{tabular}%
}
\end{table}

\begin{table*}[t!]

  \caption{\textbf{Performance comparison between different components on VISOR (\%) and Object Accuracy (OA) (\%) metrics} and T2I-score, based on Stable Diffusion 1.4 and 2.1 for different components.} 
  \centering
  \small
\resizebox{0.85\linewidth}{!}{
\begin{tabular}{cccrrrrrrr}
\toprule
 \multirow{2}{*}{$\mathcal{L}_{\text{spatial}}$}&\multirow{2}{*}{$\mathcal{L}_{\text{presence}}$}&\multirow{2}{*}{$\mathcal{L}_{\text{balance}}$}&\multirow{2}{*}{OA (\%)}&\multicolumn{6}{c}{\textbf{VISOR (\%)}}\\
\cmidrule(lr){5-10}
&&&&uncond&cond&1&2&3&4\\
\midrule

\rowcolor{gray!25}\multicolumn{3}{l}{Stable Diffusion 1.4}&\multicolumn{7}{l}{}\\

\ding{55}&\ding{55}&\ding{55} & 32.74 & 20.38& 62.25  & 49.14 & 22.07 & 8.30 & 2.01 \\
\ding{55}&\ding{55}&\ding{51} & 33.47 & 21.10& 63.03  & 50.34 & 23.26 & 8.79 & 1.99 \\
\ding{55}&\ding{51}&\ding{55} & 53.54& 31.78 & 59.36  & 64.93 & 37.75 & 18.35 & 6.09 \\
\ding{55}&\ding{51}&\ding{51} & 54.35 & 32.46& 59.73  & 65.93 & 38.72 & 18.84 & 6.37 \\
\ding{51}&\ding{55}&\ding{55} & 59.33& 58.78 & 99.07  & 87.05 & 70.83 & 50.60 & 26.65 \\
\ding{51}&\ding{55}&\ding{51} & 60.09 & 59.55 & 99.09 & 87.54 & 71.83 & 51.69 & 27.13 \\
\ding{51}&\ding{51}&\ding{55} & 66.44& 65.63 & 98.78  & 89.69 & 77.18 & 60.01 & 35.62 \\
\midrule
\rowcolor[HTML]{E7F5F8}\ding{51}&\ding{51}&\ding{51} &\textbf{ 67.36} & \textbf{66.54}& \textbf{98.79 } & \textbf{90.48} & \textbf{77.79} & \textbf{61.37} & \textbf{36.54} \\
\midrule
\rowcolor{gray!25}\multicolumn{3}{l}{Stable Diffusion 2.1}&\multicolumn{7}{l}{}\\

\ding{55}&\ding{55}&\ding{55} & 44.37& 26.91 & 60.66  & 59.30 & 30.84 & 13.66 & 3.84 \\
\ding{55}&\ding{55}&\ding{51} & 43.32 & 27.21 & 62.82 & 60.56 & 31.31 & 13.18 & 3.79 \\
\ding{55}&\ding{51}&\ding{55} & 63.09& 37.17 & 58.91  & 71.65 & 44.98 & 23.38 & 8.67 \\
\ding{55}&\ding{51}&\ding{51} & 62.85& 37.69 & 59.97  & 72.98 & 45.92 & 23.41 & 8.45 \\
\ding{51}&\ding{55}&\ding{55} & 70.58& 69.70 & \textbf{98.76}  & 92.65 & 81.53 & 65.10 & 39.53 \\
\ding{51}&\ding{55}&\ding{51} & 71.12& 70.18 & 98.69  & 92.90 & 82.20 & 65.76 & 39.86 \\
\ding{51}&\ding{51}&\ding{55} & 76.20 & 75.21 & 98.71 & 94.12 & 85.81 & 72.33 & 48.59 \\
\midrule
\rowcolor[HTML]{E7F5F8}\ding{51}&\ding{51}&\ding{51} & \textbf{77.28}& \textbf{76.26}  & 98.68 &\textbf{ 94.65} &\textbf{ 86.66} & \textbf{73.48} & \textbf{50.23} \\
\midrule

\rowcolor{gray!25}\multicolumn{3}{l}{Stable Diffusion XL}&\multicolumn{7}{l}{}\\

\ding{55}&\ding{55}&\ding{55} & 68.06 & 45.10 & 66.27 & 77.48 & 55.69 & 33.27 & 13.96\\
\ding{55}&\ding{55}&\ding{51} & 68.79 & 46.20 & 67.16 & 78.64 & 57.15 & 34.52 & 14.47\\
\ding{55}&\ding{51}&\ding{55} &     - &     - &     - &     - &     - &     - &     -\\
\ding{55}&\ding{51}&\ding{51} & 70.04 & 46.82 & 66.84 & 79.34 & 57.87 & 35.40 & 14.66\\
\ding{51}&\ding{55}&\ding{55} & 79.47 & 76.91 & 96.78 & 92.93 & 84.76 & 73.88 & 56.07\\
\ding{51}&\ding{55}&\ding{51} & 80.37 & 77.89 & \textbf{96.92} & 93.22 & 85.44 & 75.25 & 57.66\\
\ding{51}&\ding{51}&\ding{55} & 79.83 & 77.24 & 96.75 & 93.01 & 84.89 & 74.24 & 56.80\\
\midrule
\rowcolor[HTML]{E7F5F8}\ding{51}&\ding{51}&\ding{51} &\bf{81.01}&\bf{78.44}&96.83&\bf{93.64}&\bf{85.91}&\bf{75.70}&\bf{58.50}\\
\bottomrule

\end{tabular}
}

  \label{tab:losscomp_appendix}
\end{table*}

\section{Extra Qualitative Results}

\autoref{fig:qualitative_success} showcases the better generation power of \ourmethod{} compared to the Stable Diffusion baseline. The spatial losses successfully overcome the limitations in the baseline - incorrect spatial placement and single object generation. The base model generated results where objects vary in spatial location, which implies the ignorance of the base SD to the spatial relationship mentioned in the prompt. In \autoref{fig:qualitative_success}, examples with spatial relations \textit{"left of"} and \textit{"right of"} generate two misaligned objects, whereas with \textit{"above"} and \textit{"below"} the model struggles to generate both objects in one image. We attribute this mostly to the unnatural combination of the objects in the prompt: e.g. "a bench \textbf{above} a cake" \cite{chefer2023attend}. Hence, the best that SD can do is to only generate object combinations that it has seen during training - it is more likely to produce a "cat" together with a "motorcycle" than a "cake" and a "bench" in one image. The missing object problem can also be explained with some of the insights discussed in A\&E \cite{chefer2023attend}, namely that it can be suppressed, mixed with the other object, entangled in the representation of the other object or subtly blended in the image.

\ourmethod{} successfully addresses the constraints that the base SD faces by introducing well-crafted spatial losses which produce a meaningful signal used to guide the underlying diffusion model through the denoising process. Our method successfully interpreted the spatial information and generated both object at locations in accordance with the spatial relationship given in the prompt. In the rare object combination case, "a bench \textbf{above} a cake", \texttt{InfSplign} successfully disentangled the concept of the "bench" from the attention map and cleverly figures out that the bench object cannot realistically be placed on top of a cake, so it generates it as a cake topper above the cake.

\begin{figure}[h!b]
\centering
\includegraphics[width=1\linewidth]{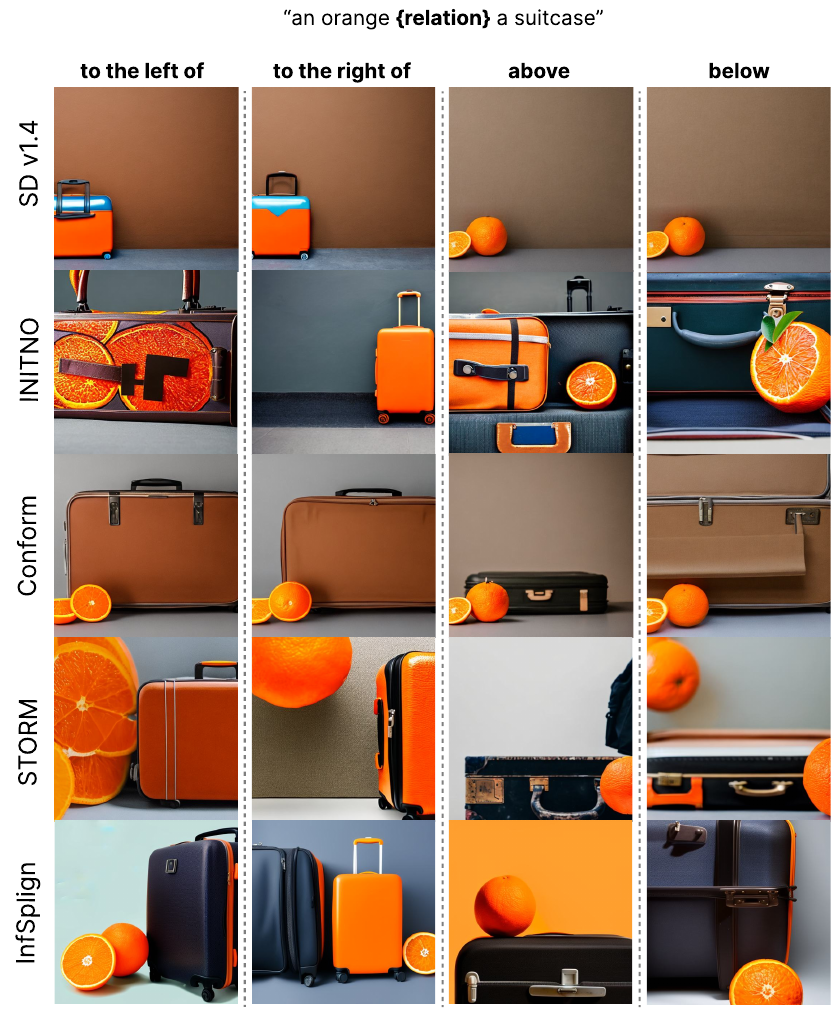}
\caption{An Orange \{relation\} a Suitcase.}
\label{fig:orange-suitcase}
\end{figure}

\begin{figure}[h!t]
\centering
\includegraphics[width=1\linewidth]{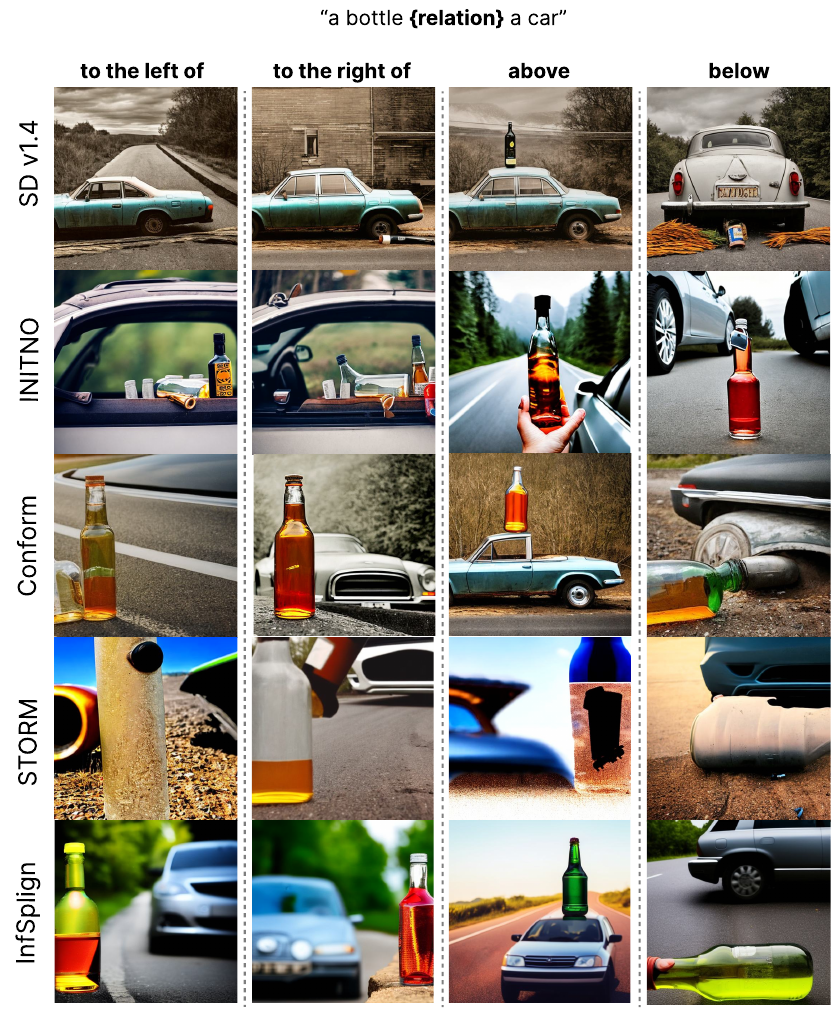}
\caption{A Bottle \{relation\} a Car.}
\label{fig:bottle-car}
\end{figure}

\begin{figure}[h!t]
\centering
\includegraphics[width=1\linewidth]{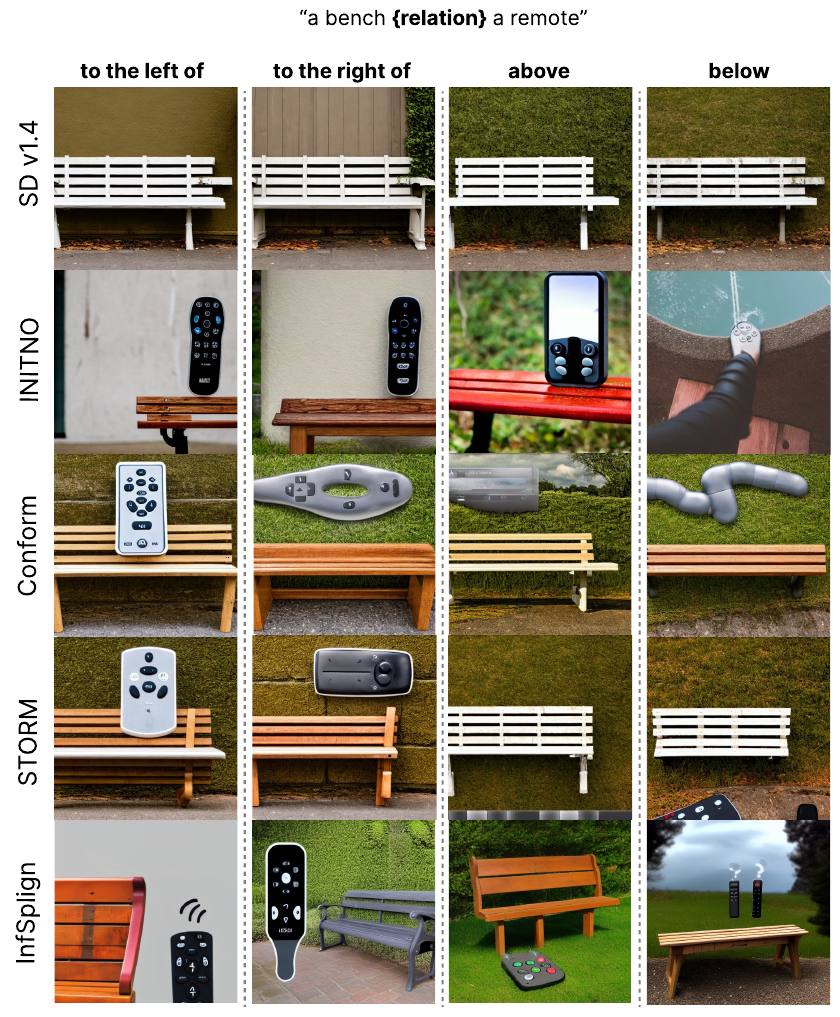}
\caption{A Bench \{relation\} a Remote.}
\label{fig:bench-remote}
\end{figure}

\begin{figure}[h!t]
\centering
\includegraphics[width=1\linewidth]{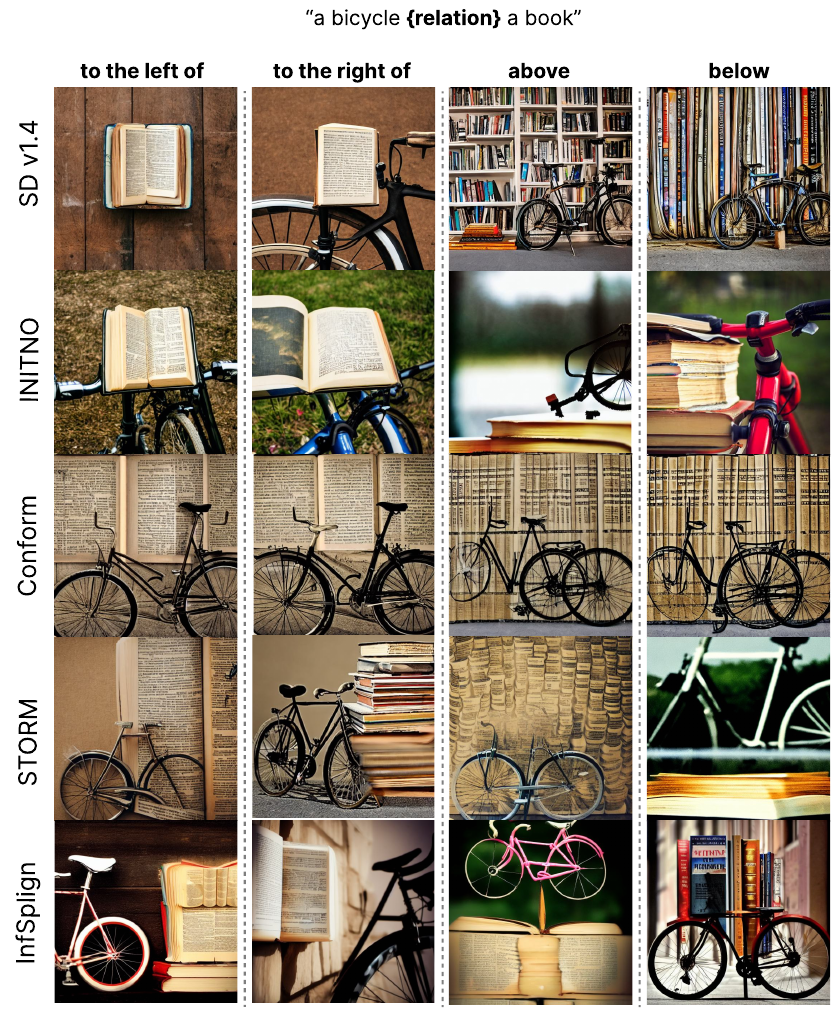}
\caption{A Bicycle \{relation\} a Book.}
\label{fig:bicycle-book}
\end{figure}

\begin{figure}[h!t]
\centering
\includegraphics[width=1\linewidth]{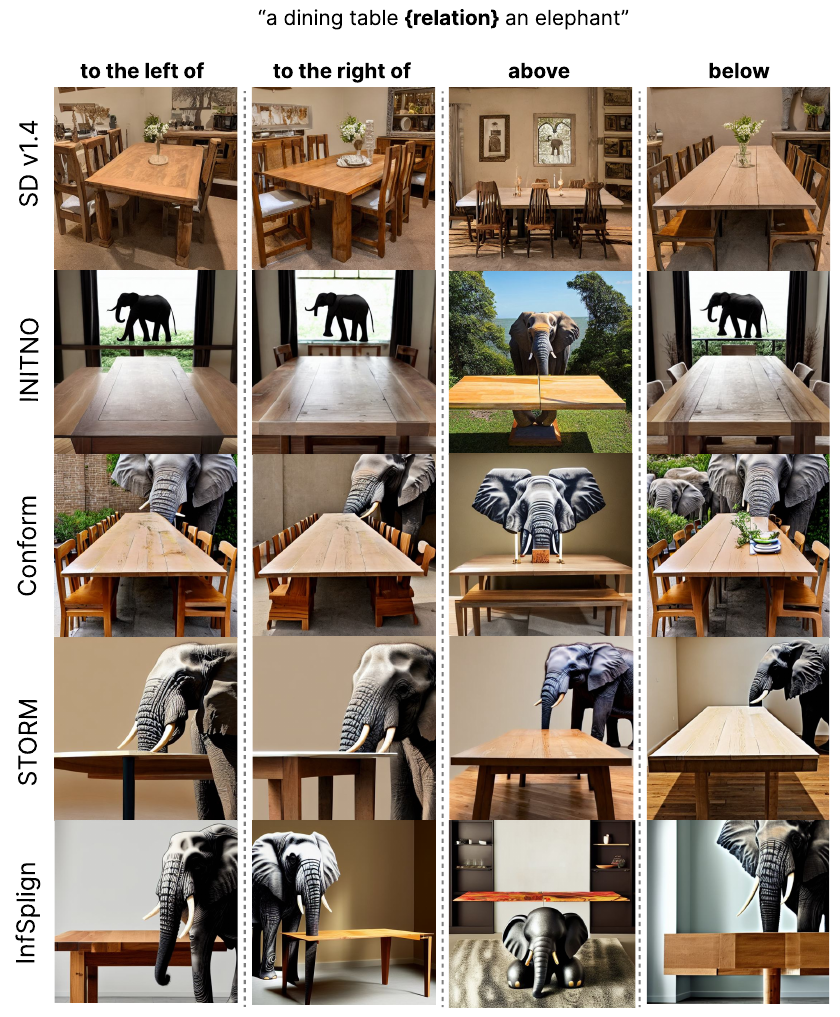}
\caption{A Dining Table \{relation\} an Elephant.}
\label{fig:dining-table-elephant}
\end{figure}

\begin{figure}[h!t]
\centering
\includegraphics[width=1\linewidth]{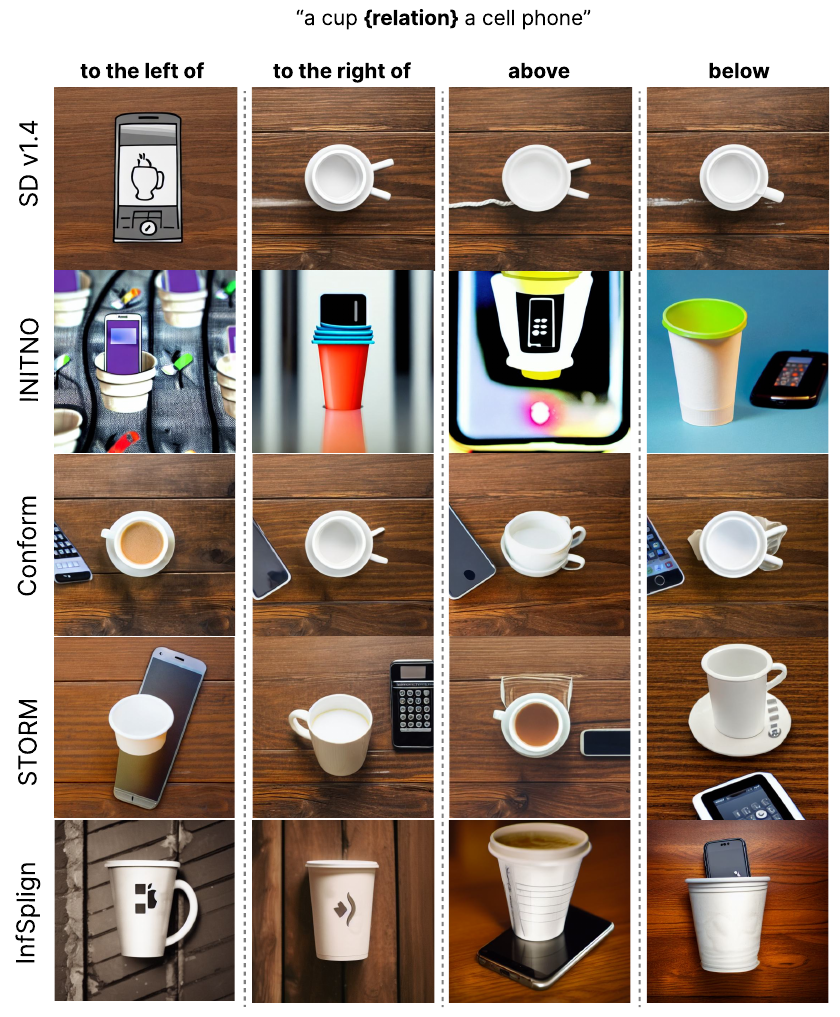}
\caption{A Cup \{relation\} a Cell Phone.}
\label{fig:cup-cellphone}
\end{figure}

\begin{figure}[h!t]
\centering
\includegraphics[width=1\linewidth]{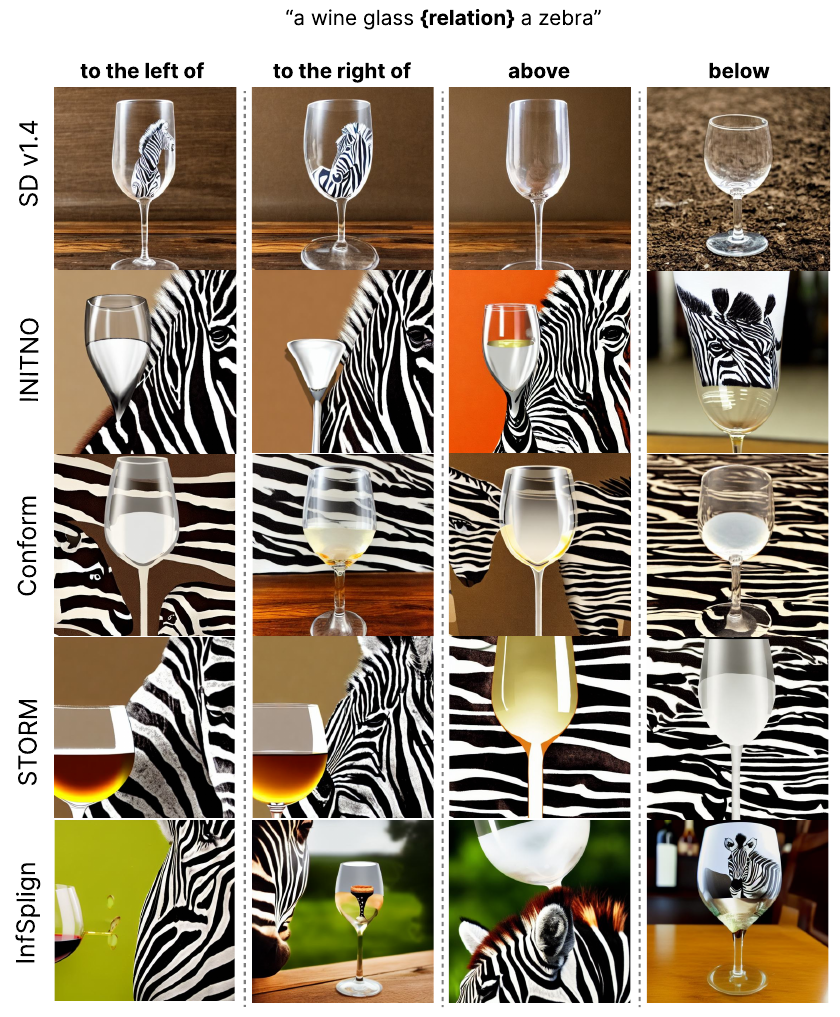}
\caption{A Wine Glass \{relation\} a Zebra.}
\label{fig:wine-glass-zebra}
\end{figure}

To further exhibit the power of \ourmethod{} to spatially align objects in the image, \autoref{fig:orange-suitcase} through \autoref{fig:wine-glass-zebra} showcase qualitative comparisons with the relevant competitors (SD v$1.4$ \cite{rombach2022high}, INITNO \cite{guo2024initno}, CONFORM \cite{meral2024conform} and STORM \cite{han2025spatial}). Note that \ourmethod{} qualitatively proves the gains showcased in the earlier mentioned quantitative analysis in \autoref{tab:visor_results} and \autoref{tab:t2i_results}.

\renewcommand{\arraystretch}{0.85}
\label{app:extra_qual_results}

\end{document}